\documentclass[10pt,twocolumn,letterpaper]{article}

\usepackage[pagenumbers]{cvpr} 

\usepackage[dvipsnames]{xcolor}

\usepackage{algorithm}
\usepackage{algorithmic}
\usepackage{graphicx}
\usepackage{siunitx}
\usepackage{amssymb}
\usepackage[colorlinks,linkcolor=blue]{hyperref}

\usepackage{times}
\usepackage{epsfig}
\usepackage{amsmath}
\usepackage{booktabs}

\definecolor{cvprblue}{rgb}{0.21,0.49,0.74}

\def\paperID{9468} 
\def\confName{CVPR}
\def\confYear{2024}

\title{GTAutoAct: An Automatic Datasets Generation Framework Based on Game Engine Redevelopment for Action Recognition}

\author{
{Xingyu Song \quad Zhan Li \quad Shi Chen \quad Kazuyuki Demachi}\\
The University of Tokyo, Japan\\
{\tt\small \{songxingyu0429, lizhan, shichen, yypr9411\}@g.ecc.u-tokyo.ac.jp}
} 

\begin{document}

\renewcommand{\thefootnote}{\fnsymbol{footnote}}
\twocolumn[{
\renewcommand\twocolumn[1][]{#1}
\maketitle
\vspace{-0.9cm}
\begin{center}
\captionsetup{type=figure}
\includegraphics[width=\textwidth]{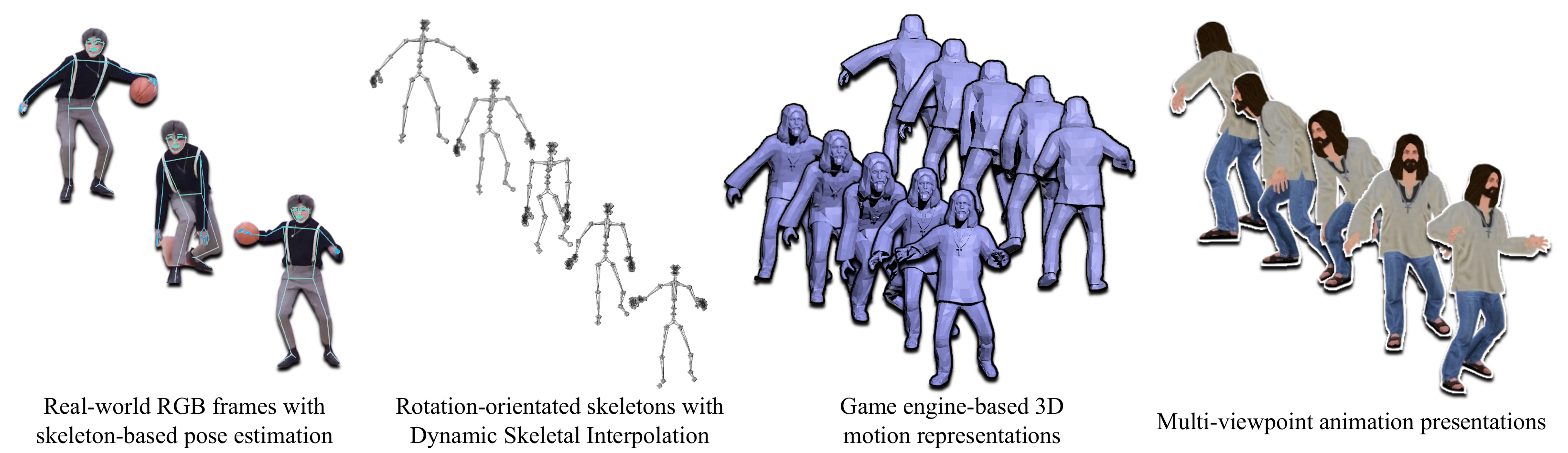} 
\captionof{figure}{Overview of GTAutoAct.}
\label{fig:overall}
\vspace{0.5cm}
\end{center}
}]

\maketitle
\begin{abstract}
\vspace{-0.5cm}
Current datasets for action recognition tasks face limitations stemming from traditional collection and generation methods, including the constrained range of action classes, absence of multi-viewpoint recordings, limited diversity, poor video quality, and labor-intensive manually collection. 
To address these challenges, we introduce GTAutoAct, a innovative dataset generation framework leveraging game engine technology to facilitate advancements in action recognition. 
GTAutoAct excels in automatically creating large-scale, well-annotated datasets with extensive action classes and superior video quality. 
Our framework's distinctive contributions encompass: 
(1) it innovatively transforms readily available coordinate-based 3D human motion into rotation-orientated representation with enhanced suitability in multiple viewpoints; 
(2) it employs dynamic segmentation and interpolation of rotation sequences to create smooth and realistic animations of action; 
(3) it offers extensively customizable animation scenes; 
(4) it implements an autonomous video capture and processing pipeline, featuring a randomly navigating camera, with auto-trimming and labeling functionalities. 
Experimental results underscore the framework's robustness and highlights its potential to significantly improve action recognition model training.
\end{abstract}    
\section{Introduction}
\label{sec:intro}

The existing datasets for action recognition task, e.g.\ Kinetics~\cite{dataset05_Kinetics}, UCF101~\cite{dataset03_UCF} and AcitvityNet~\cite{dataset04_ActivityNet}, significantly contribute to the accuracy on benchmark action recognition methods, like TSN~\cite{model01_TSN} and TSM~\cite{model02_TSM}.
These datasets play a crucial role in enhancing the understanding of human actions. 

However, existing datasets still remain shortcomings due to conventional data collection methods in the real world.
(1) Constrained range of action classes: Many datasets focus on a restricted set of actions, typically those more common or easily captured in real-world scenarios. 
This biased distribution can lead to unsatisfying performance in action recognition tasks when encountering less frequent or more specialized situations. 
It can also impact the evaluation of novel recognition models, offering a less comprehensive understanding on their capabilities. 
(2) Absence of multi-viewpoint recordings: Capturing a specific action from multiple angles is often unfeasible in real-world settings. 
Consequently, the understanding and analysis of such actions are limited to a few isolated viewpoints, rather than encompassing their entire spectrum of movement.
(3) Limited diversity: Real-world datasets are typically limited in a restricted range of environments and characters, which constrains the model's performance in recognizing actions across varied settings.
(4) Poor video quality: Videos recording from the early years suffer from the poor recording equipment or video processing technique, which can cause trouble for identifying human, generating optical flow or estimating human keypoint skeleton~\cite{dataset13_finegym}. 
(5) Labor-intensive manually collection: The process of manually collecting and annotating datasets demands considerable time and human labor, resulting in increased costs.


To address these problems, we introduce GTAutoAct, an innovative dataset generation framework leveraging game engine technology to facilitate advancements in action recognition. 
GTAutoAct is capable of automatically generating large-scale datasets for any type of action, complete with full annotations for action recognition tasks. 
It offers multiple perspectives and fully customized scenarios, with superior resolution and frame rate.
The overview of GTAutoAct framework is depicted in Figure~\ref{fig:overall}.

Furthermore, we conduct experiments to assess the ability of GTAutoAct in bridging the domain gap between virtual representations and real-world tasks.
This evaluation involved testing the datasets generated by GTAutoAct for both whole-body and major-part motion representations.

\section{Related Work}
\label{sec:formatting}

{\bf Action recognition datasets.} 
In recent years, numerous exceptional datasets for action recognition tasks have been developed. 
For instance, HMDB51~\cite{dataset02_HMDB} is noted for its well-organized annotations; 
Kinetics-[400/600/700]~\cite{dataset15_kin700,dataset05_Kinetics,dataset14_kin600} are recognized for their variety of classes.
UCF101~\cite{dataset03_UCF} stands out for its detailed classification.
Moments in Time and Multi-Moments in Time~\cite{dataset10_mit,dataset11_mmit} are known for their extensive collection of action clips.
HVU~\cite{dataset01_HVU} is distinguished for its detailed annotation encompassing actions, attributes, and concepts.
FineGym~\cite{dataset13_finegym} offers datasets that are both fine-grained and coarse-grained. 
However, none of these datasets features automatic data generation and collection capabilities.

{\bf Video game-based action datasets.} 
Video game-based datasets have been increasingly utilized for training deep learning models. 
JTA~\cite{jta}, for example, is a vast dataset created using a video game for pedestrian pose estimation and tracking in urban environments. 
GTA-IM~\cite{gamedataset02_GTAIM} a pose estimation dataset, highlights human-scene interactions and employs a developed game engine interface for automatic control of characters, cameras, and actions. 
However, both them focus on static human poses, rather than capturing temporal movements. 
NCTU-GTA360~\cite{gamedataset04_GTA360}, an action recognition dataset featuring spherical projection captured from video games, involves the use of 360-degree cameras to record the entire surroundings of a character.
Nevertheless, NCTU-GTA360 faces an imbalance in action distribution, with basic actions like "OnFoot" or "Stopped" overwhelmingly dominating more complex actions such as "Ragdoll" or "SwimmingUnderWater." 
Other datasets like SIM4ACTION~\cite{gamedataset03_SIM} and G3D~\cite{gamedataset01_G3D} also share a common limitation in offering a constrained range of action classes. 
Most notably, none of these datasets, including GTA-IM and NCTU-GTA360, have succeeded in effectively importing a vast amount of self-customized actions from the real world, which is crucial for creating more comprehensive and diverse training models.

{\bf Synthetic data generation.} 
Several methods have been devised for generating synthetic data tailored to action recognition tasks.
ElderSim~\cite{ElderSim}, for instance, is a simulation platform designed for creating synthetic data pertaining to the daily activities of elderly individuals.
It offers the capability to generate realistic motions of synthetic characters, providing a range of adjustable data-generating options.  
However, ElderSim's utility is somewhat limited as it is not designed for general action generation across various categories.
SURREACT~\cite{synthetic_humans} involves the use of monocular 3D human body reconstruction to render synthetic training videos for action recognition, which also demonstrates that synthetic humans can enhance the performance of action recognition tasks. 
However, a significant drawback of this method is the undeniable mismatch between characters and surrounding environments, with actions lacking in smoothness and photorealism. 
These issues can potentially harm the effective training of action recognition models.

\section{GTAutoAct}
\label{sec:gtautoact}

GTAutoAct\footnotemark[1]\footnotetext[1]{For more information, please refer to the appendix.} is a framework designed to automatically generate datasets for action recognition tasks. GTAutoAct encompasses three integral modules: Action animation, Scene customization, and Auto-collection. 

\subsection{Action Animation}
The Action Animation module is designed to capture real-world actions from a singular viewpoint, subsequently transforming them into animations viewable from a plenary viewpoint, covering all possible angles, for presentation within the gaming environment.

An animation of action is consisted by a sequence of static but continuous poses across each frame. Each pose is intricately defined by the 3D human skeleton, constituted by an interconnected network of joint nodes and edges that span between them. Differ from conventional approach, which utilizing the 3D coordinates of each joint node to represent a pose, our methodology defines each bone joint, comprising two adjacent nodes and the edge connecting them, as a discrete motion unit. By capturing the changes of rotation from its initial to its current position within a pose, we achieve a precise representation of poses. 

There are primarily two advantages resulting from the utilization of rotations instead of coordinates in our methodology: 
(1) Enhanced suitability for diverse characters. Simply relying on coordinates for pose representation leads to severe and inevitable mismatches due to the extensive range of variations in height, body shape, age and gender among different characters. Conversely, their rotations of bones, which describe the relative angles between body parts, remain consistent across each instance if the characters are positioned in an identical pose.
(2) More intuitive manipulation and realistic constrains. Rotations provide a more straightforward way to manipulate body motion, by directly adjusting the angle of a bone joint, which can be more efficient than manipulating coordinates. Additionally, this feature of rotation can lead to more realistic constraints on movements, preventing unnatural poses and jitter issues. 
(3) Simplified interpolation and data smoothing. Rotations contain more sophisticated motion information compared to coordinates, thereby avoiding movement distortion or vanishing that might occur with frame interpolation based on coordinates.

\vspace{-0.1cm}
\subsubsection{Rotation-orientated 3D human motion representation system}
\begin{figure}[t]
\centering
\includegraphics[width=\columnwidth]{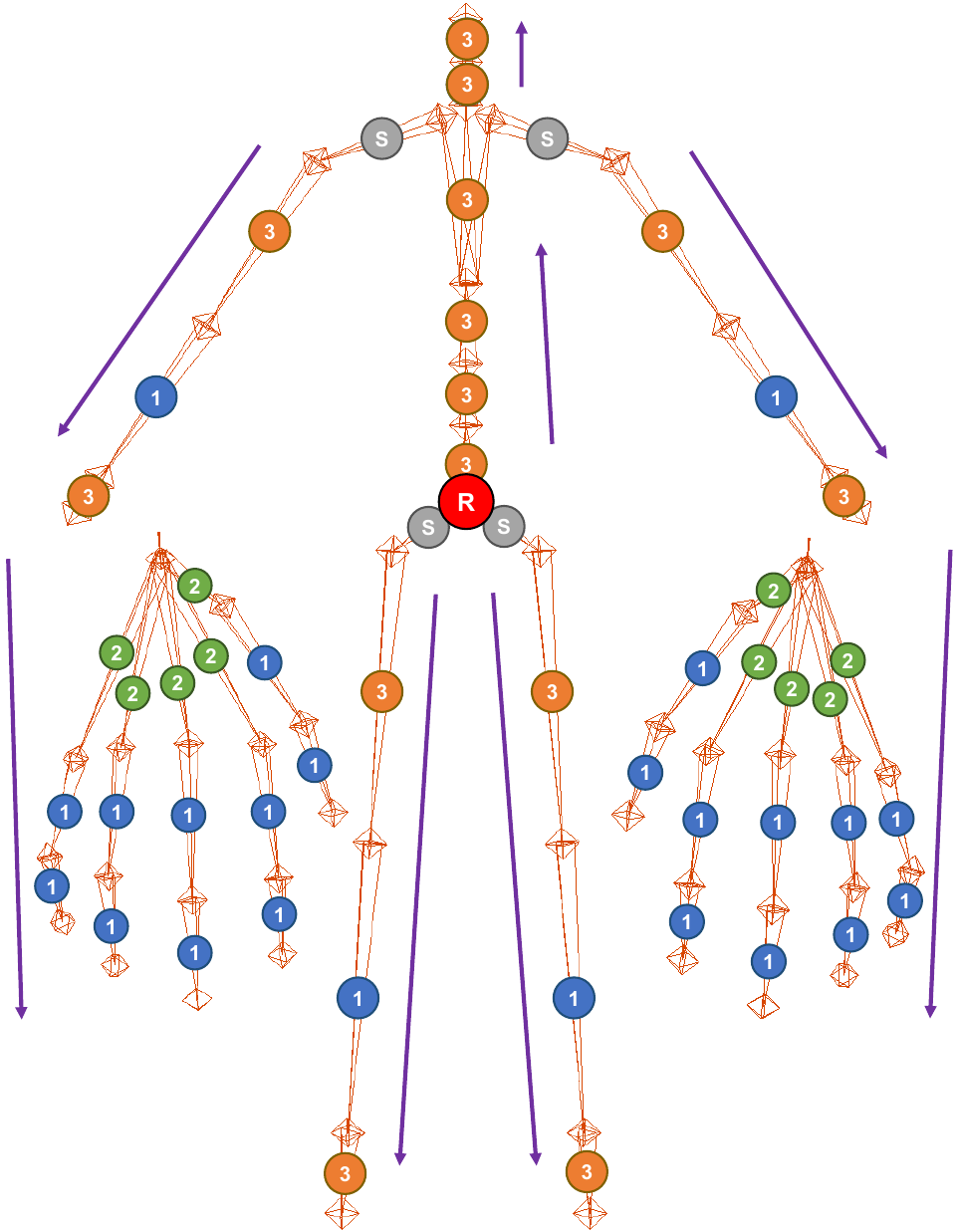} 
\caption{Configuration of 53 bone joints in human motion representation system in GTAutoAct. Red sign labeled with ``R'' denotes the root bone joint. Gray signs labeled with ``S'' denote the static bone joints. Orange, green, and blue signs labeled with ``3'', ``2'', ``1'' respectively, correspond to three-, two-, and one-dimensional bone joints. Violet arrows denote the hierarchical inheritance directions from parent joint to child joint.  }
\vspace{-0.3cm}
\label{configuration}
\end{figure}
First of all, we configured a system with 53 bone joints to represent whole-body human motions. 
The specific configuration is detailed in Figure~\ref{configuration}. 
Within this system, all bone joints can be categorically divided into four distinct types based on their rotational degree of freedom, three-, two-, one-dimensional, and static bone joints. 
Three-dimensional bone joints, such as shoulders and spine, possess rotational degrees of freedom along three mutually perpendicular axes: yaw, pitch, and roll\footnotemark[1].
Conversely, two-dimensional bone joints, such as metacarpal bones which facilitate the opening and closing of hands, lack rotational freedom around the axis of the bone itself. 
In the case of one-dimensional bone joints, such as the forearms and calves, which are controlled by the elbows and knees respectively, their rotations are constrained to a single dimension, specifically around the pitch axis. 
In addition, the root bone joint can be regarded as a special one-dimensional bone joints, processing exclusive rotational freedom on a horizontal plane ($xOy$ plane), as it functions to indicate the overall facing direction of the whole body throughout an action. 
In cases where bone joints exhibit negligible rotational variations throughout an action, we categorize them as static bone joints to simplify the computational process. Examples of such joints include the clavicles and pelvis. 

Furthermore, the rotation of each bone joint is articulated with respect to the Local Coordinate System (also known as Parent Coordinate System), ensuring that the rotation angles of a specific bone joint are defined exclusively in relation to its parent node. 
Starting from the root bone joint, the upper and lower body respectively form an hierarchical inheritance structure, extending from the parent node to the child node of each joint.

\subsubsection{Coordinate transformation}

We employ coordinate transformation to convert the 3D coordinates of discrete joint nodes within the human body skeleton into the rotations of each bone joint in our motion representation system. Additionally, we articulate these bone joints using 3D vectors.
For example, the vector representing the upper arm is derived by subtracting the coordinates of the shoulder node from those of the elbow node: 
\begin{equation}
\Vec{v}=\boldsymbol{p}_{elbow}-\boldsymbol{p}_{shoulder}
\end{equation}

\begin{figure}[t]
\centering
\includegraphics[width=1\columnwidth]{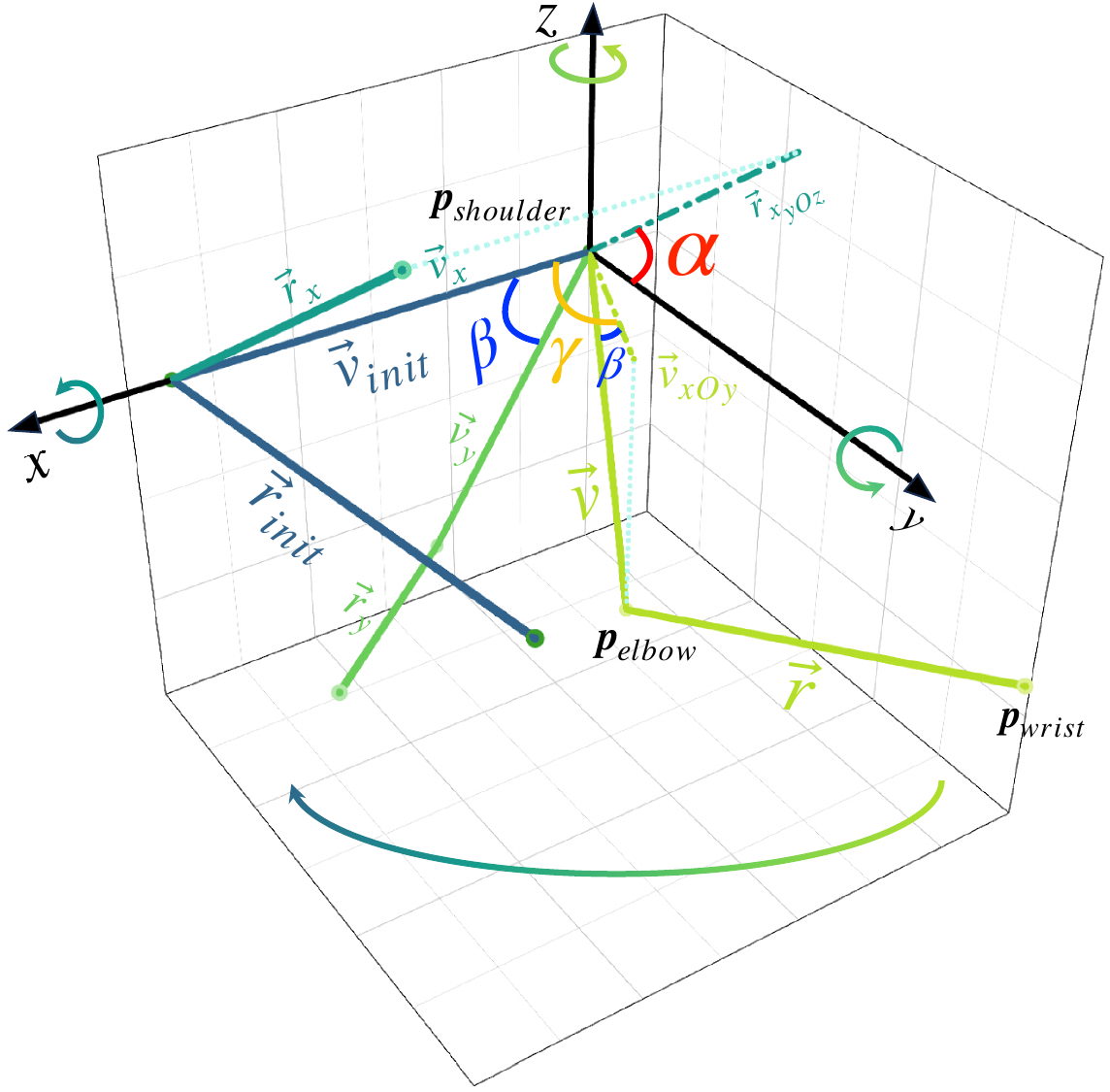} 
\caption{Schematic diagram of Euler angle calculation. 
$\boldsymbol{p}_{shoulder}$ (rotation center), $\boldsymbol{p}_{elbow}$, and $\boldsymbol{p}_{wrist}$ represent the nodes corresponding to the shoulder, elbow and wrist joints in 3D World Coordinate System respectively. 
Given the initial target vector $\Vec{v}_{init}$ and initial reference vector $\Vec{r}_{init}$, and proceeding to rotate them sequentially around the $x,y \text{ and } z$ axes by angles $\alpha,\beta, \text{ and } \gamma$ respectively, $(\Vec{v}_x,\Vec{r}_x)$, $(\Vec{v}_y,\Vec{r}_y)$ and finally $(\Vec{v},\Vec{r})$ can be obtained. 
The diagram displays a gradient arrow, signifying the Euler angle calculation order, which is in reverse to the rotation sequence.
Consequently, we calculate $\gamma$ as the rotation angle from the positive $x$-axis to $\vec{v}_{xOy}$, which representing the projection of vector $\vec{v}$ onto the $xOy$ plane. 
Likewise, $\beta$ is inferred as the rotation angle between vector $\vec{v}$ and the $xOy$ plane. 
Lastly, $\alpha$ is calculated as the angle spanning from the positive $y$-axis to $\vec{r}_{x_{yOz}}$, the projection of $\vec{r}_{x}$ onto the $yOz$ plane.}
\vspace{-0.2cm}
\label{euler_angle}
\end{figure}

Therefore, poses can be articulated by the rotations of each bone vector from the initial position to the current position. 
Initially, We compute the static Euler angles\footnotemark[1] of each vector in World Coordinate System in $x-y-z$ sequence, utilizing the corresponding coordinates of skeleton nodes. 
To illustrate, Figure~\ref{euler_angle} demonstrates the calculation of the Euler angles $(\alpha,\beta,\gamma)$ for the upper arm vector $\Vec{v}$, with shoulder joint node $\boldsymbol{p}_{shoulder}$ as the center of rotation. 
Hence, Euler angles $\gamma$ and $\beta$ can be calculated as:
\begin{equation}
    \vec{v}_{xOy}=\vec{v}-\langle \vec{v},\vec{n}_{z} \rangle 
\end{equation}
\begin{equation}
    \gamma=\arccos{
    \frac{\vec{v}_{xOy} \cdot \vec{i}_{x}}
    {||\vec{v}_{xOy}||}    
}
\end{equation}
\begin{equation}
    \beta=\arccos{
    \frac{\vec{v}_{xOy} \cdot \vec{v}}
    {||\vec{v}_{xOy}|| \cdot ||\vec{v}||}   
    }
\end{equation}
where $\vec{n}_{z}$ represents the normal vector of $xOy$ plane and $\vec{i}_{x}$ signifies the unit vector along $x$-axis. 

Nonetheless, when a 3D vector is parallel to the $x$-axis, it becomes incapable of representing a rotation about the $x$-axis.
To address this, we introduce a reference vector\footnotemark[1] $\vec{r}$, represented by the vector formed by the child bone joint of the target bone joint, such as the forearm (child) and upper arm (parent), aiding in the calculation of $\alpha$. 
Consequently, $\alpha$ can be calculated using $\vec{r}$, $\gamma$, and $\beta$ as follows:

\begin{equation}
    \vec{r}_{x}=RM^{-1}_{z}(\gamma) \cdot RM^{-1}_{y}(\beta) \cdot \vec{r}
\end{equation}
\begin{equation}
    \alpha=\arccos{
    \frac{(\vec{r}_{x}-\langle \vec{r}_{x},\vec{n}_{x} \rangle ) \cdot \vec{i}_{y}}
    {||\vec{r}_{x}-\langle \vec{r}_{x},\vec{n}_{x} \rangle ||}    
}
\end{equation}
where $\vec{r}_{x}$ represents the vector $\vec{r}$ rotated around $x$-axis by angle $\alpha$. $RM^{-1}_{z}(\gamma)$ and $RM^{-1}_{y}(\beta)$ denote the inverse rotation metrics for angles $\gamma$ and $\beta$ respectively. $\vec{n}_{x}$ denotes the normal vector to $yOz$ plane, and $\vec{i}_{y}$ is the unit vector along $y$-axis.

However, to avoid the potential Gimbal Lock issue associated with Euler angles, we convert these angles into quaternions\footnotemark[1] for rotation representation. 
The transformation from Euler angles to quaternions is illustrated below:
\begin{equation}
    \boldsymbol{q}=q_{x}\boldsymbol{i}+q_{y}\boldsymbol{j}+q_{z}\boldsymbol{k}+q_{w}
\end{equation}
\begin{equation}
    q_{x}=\sin{(\theta/2)}\cdot \cos{\alpha} 
\end{equation}
\begin{equation}
    q_{y}=\sin{(\theta/2)}\cdot \cos{\beta} 
\end{equation}
\begin{equation}
    q_{z}=\sin{(\theta/2)}\cdot \cos{\gamma} 
\end{equation}
\begin{equation}
    q_{w}=\cos{(\theta/2)} 
\end{equation}
where $\theta$ represents the rotation angle transitioning from the initial position to the current position\footnotemark[1], also serves as the composition of rotations by angles $\alpha$, $\beta$ and $\gamma$ in 3D coordinate system.

Accordingly, the coordinates and quaternions in Local (Parent) Coordinate System can be calculated as follows:
\begin{equation}
   TM_{parent}= 
    \left(
        \begin{array}{cc}
            RM_{parent} & \boldsymbol{p}_{parent} \\
            \boldsymbol{0} & 1
        \end{array}
    \right)
\end{equation}
\begin{equation}
    \boldsymbol{p}_{|WCS} = [x,y,z,1]^{\mathsf{T}}
\end{equation}
\begin{equation}
    \boldsymbol{p}_{|LCS}=TM^{-1} \cdot \boldsymbol{p}_{|WCS}
\end{equation}
\begin{equation}
    \boldsymbol{q}_{child|LCS} = Inv(\boldsymbol{q}_{parent|WCS}) \times \boldsymbol{q}_{child|WCS}
\end{equation}
where $TM_{parent}$ is a $4\times 4$ transformation matrix\footnotemark[1] that combines rotation and translation components, with $RM_{parent}$ being the rotation matrix of the parent node, and $\boldsymbol{p}_{parent}$ being a $3\times1$ array, representing the coordinates of the parent node.
$\boldsymbol{p}_{|LCS}$ and $\boldsymbol{p}_{|WCS}$ represent the $4 \times 1$ homogeneous coordinates of a point $p$, in the Local and World Coordinate Systems respectively.
$\boldsymbol{q}_{i|S}$ represents the quaternion of vector $\vec{v}_{i}$ in coordinate system $S$. 
$Inv(\boldsymbol{q})$ denotes the inverse of quaternion $\boldsymbol{q}$, while ``$\times$'' denotes the quaternion multiplication.

Ultimately, human pose can be represented by quaternion set $Q$ and root coordinates $\boldsymbol{p}_{root}$, expressed as follows:
\begin{equation}
\boldsymbol{R}=\mathcal{M}(Q\,, \boldsymbol{p}_{root})\,, Q=\{ \boldsymbol{q}_{b|LCS} \mid b \in B\}
\end{equation}
where $\mathcal{M}(\cdot)$ denotes the mesh generation process\footnotemark[1] conducted by game engine, and $B$ denotes the set of bone joints. 
\subsubsection{Dynamic skeletal interpolation}

After assembling the rotations of each bone joint for an individual frame, we proceed to compile them to create a continuous rotation sequence across a stream of frames to depict a complete representation of action (so-called animation). 
However, the unsatisfying results in the 3D human pose estimation task may lead to frequent mismatches and jitters among the animations. 
Unlike typical time series data, directly applying a piecewise polynomial interpolation algorithm to rotation sequences can lead to significant issues, such as missing specific poses and decreased movement amplitude. 
To tackle these challenges, we have developed the Dynamic Skeletal Interpolation algorithm\footnotemark[1]. 
This algorithm dynamically segments the rotation sequence into unit motions based on the variation in the range of motion. 
It then automatically interpolates the rotation sequence across different frame counts, ensuring a smoother and more natural-looking animation.
Finally, it randomly generates a series of variants for each animation, ensuring a diverse representation of each action.

The algorithm is detailed in Algorithm~\ref{alg}. 
Here, $d^{f}$ denoted the weighted average of the Angular Distance of each bone joint between adjacent frames, with $w_{b}$ representing the weight of each bone joint. 
The functions $Re(\cdot)$ and $conj(\cdot)$ return the real part and the conjugate of a quaternion respectively. 
$p(x)_{[a,b]}$ refers to the Lagrange interpolating polynomial in the interval $[a,b]$, while $L(x)$ represents the Lagrange basis polynomial. 
$Linespace([a,b],n)$ is the function used to create evenly space numbers over interval $[a,b]$. 
The parameter $\delta$ is the interpolation rate, indicting the ratio of original frames to interpolated frames. 
$\eta$ is the interpolation coefficient.  
To enhance the diversity of an action, we develop Random Variation function $\mathcal{V}(\cdot)$ to generate a series of variants for each animation.
To smooth the piecewise interpolated data, we employ Supersmoother~\cite{supersmoother} $\mathcal{S}(\cdot)$, a non-parametric smoothing method. 
$V, J, F, F'$ denote the number of variants, bone joints, frames before and after interpolation.

\begin{algorithm}[tb]
\caption{Dynamic Skeletal Interpolation}
\label{DSI}
\textbf{Input}: $\boldsymbol{A} \in \mathbb{R}^{J \times F \times 4}$
\begin{algorithmic}[1] 
\STATE Define $\boldsymbol{A}' \in \mathbb{R}^{J \times F' \times 4}$
\FOR{$f=2$ to $F$}
    \STATE $d^{f} = \frac{1}{J}\displaystyle\sum_{b \in B} w_{b} \cdot 2\arccos{(Re(\boldsymbol{q}_{b}^{f} \times conj(\boldsymbol{q}_{b}^{f-1}))}$
    \IF{$d^{f}>threshold$}
        \STATE $p_{[i,f-1]}(x) = \displaystyle\sum_{b \in B} \sum_{t=i}^{f-1} \boldsymbol{q}_{b}^{t} \cdot L_{t}(x)$
        \STATE $p_{[f-1,f]}(x) = \displaystyle\sum_{b \in B} 
        (\boldsymbol{q}_{b}^{f-1} \cdot L_{f-1}(x) + \boldsymbol{q}_{b}^{f} \cdot L_{f}(x))$
        \STATE $\boldsymbol{x}_{nor} = Linespace([i,f-1], \frac{1}{\delta})$
        \STATE $\boldsymbol{x}_{edge} = Linespace([f-1,f], Int(\frac{\eta \cdot d^{f}}{\delta}))$
        \STATE $\boldsymbol{A}'_{[i,f-1]}=p_{[i,f-1]}(\boldsymbol{x}_{nor})$
        \STATE $\boldsymbol{A}'_{[f-1,f]}=p_{[f-1,f]}(\boldsymbol{x}_{edge})$
        \STATE $i = f-1$
    \ENDIF
\ENDFOR
\STATE $\boldsymbol{\Phi} = Supersmoother(\mathcal{V}(\boldsymbol{A}',V))$
\end{algorithmic}
\textbf{Output}: $\boldsymbol{\Phi}\in \mathbb{R}^{V\times J \times F' \times 4}$
\label{alg}
\end{algorithm}

\subsection{Scene Customization}

\begin{figure}[t]
\centering
\includegraphics[width=\columnwidth]{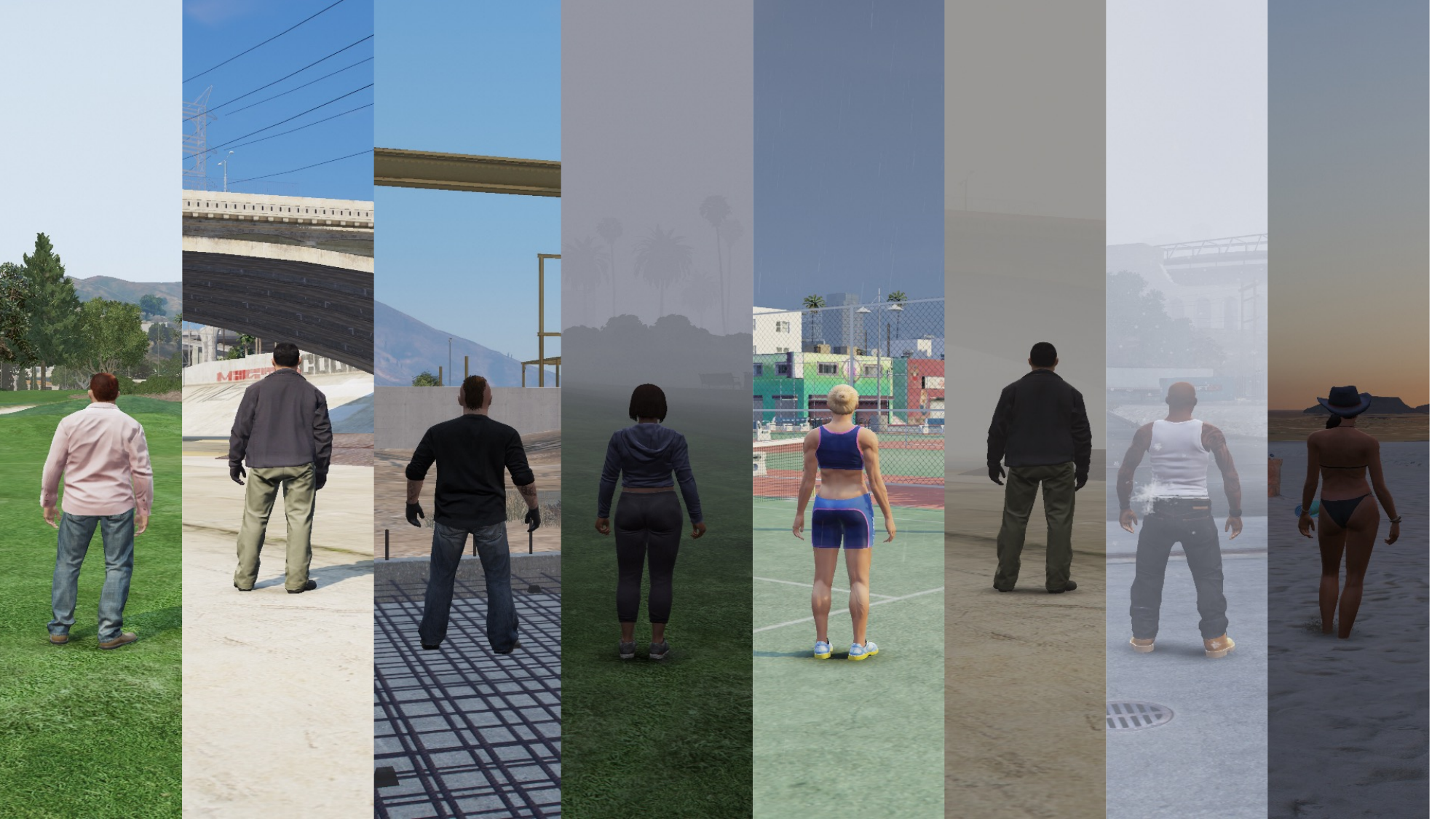} 
\caption{Examples of customized scenes in FiveM.}
\vspace{-0.3cm}
\label{scene_customization}
\end{figure}

The action animations are showcased using FiveM~\cite{fivem}, a modification platform for GTAV, enabling players to play multi-players on customized dedicated server. 
Scene Customization\footnotemark[1] aims to craft scenes for these action animations in FiveM. 
The examples of customized scenes are illustrated in Figure~\ref{scene_customization}. 
Scene Customization encompasses three main aspects: Environmental customization for setting specific environmental situations; Character customization to modify characters for various scenarios; and Map editing to tailor the landscapes. 
These adjustments accommodate unique action situations and enhanced data diversity.

Environmental customization involves altering weather conditions, time of day, and in-game locations. 
Through FiveM scripts, we facilitate random weather variations, time adjustments, and repositioning across different in-game locales to achieve varied scenarios.
Character customization facilitates modifying both native FiveM characters and player-created avatars, enabling the simulation of actions by individuals in diverse ages, genders, and professions, reflecting the variety found in real-world scenarios.

Map customization enables editing of the in-game landscape and the attributes of entities, which also supports the integration of player-created custom entities.

\subsection{Auto-collection}
Auto-collection focuses on collecting videos of in-game animations and trimming them automatically. 

\subsubsection{Random Camera Moving algorithm}
\begin{figure}
\begin{center}
\centering
\includegraphics[width=\linewidth]{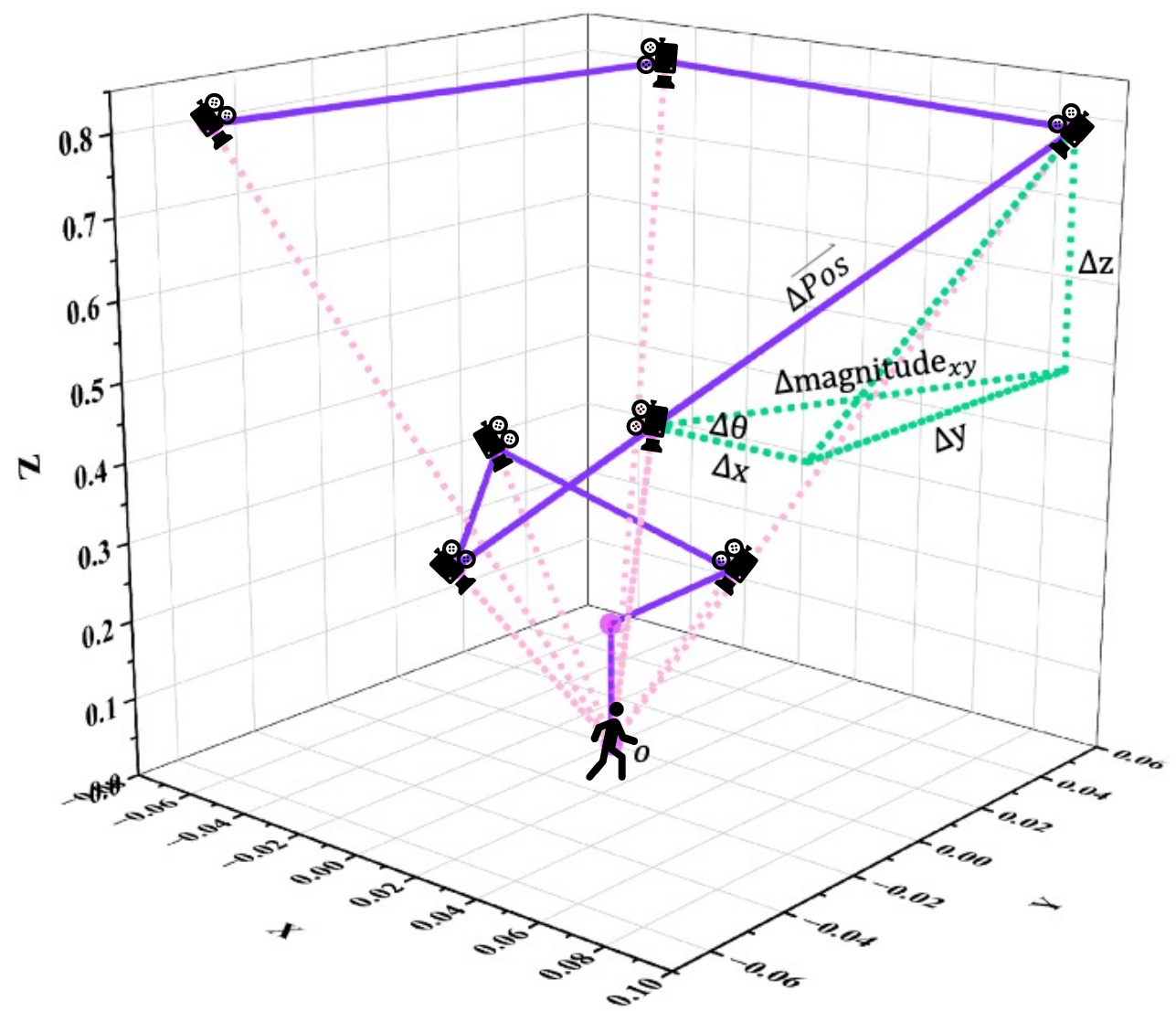}
\end{center}
\caption{Schematic diagram of RCM. The camera-shaped icons are the camera positions derived by RCM. The purple lines are the movement of camera between adjacent positions. 
$o$ is the initial point of camera and the position of character as well. 
$\Delta{magnitude_{xy}}$ and $\Delta\theta$ indicate the random distance and angle change in horizontal plane $xOy$. 
$\Delta{z}$ indicates the random height change in vertical $z$-axis. }
\vspace{-0.2cm}
\label{rcm}
\end{figure}

The Random Camera Moving (RCM) algorithm navigates the camera to capture a specific action animation from varied viewpoints, as shown in Figure~\ref{rcm}. 
According to RCM algorithm, the camera starts at the character’s position, considered the origin (point $o$).
For each move, the camera traverses a random horizontal distance at a random angle within the horizontal plane, and additionally shifts a random vertical distance. 
Therefore, the transition between two consecutive points $Pos_{i}$ and $Pos_{i+1}$ can be represented as follows:
\begin{equation}
\overrightarrow{Pos}_{i+1}=\overrightarrow{Pos}_{i}+\Delta\overrightarrow{Pos}
\end{equation}

Where $\Delta\overrightarrow{Pos}$ denotes the change of the position vector, which can be expressed as:

\begin{equation}
\Delta\overrightarrow{Pos}_{xy}=\Delta\overrightarrow{Pos}-\Delta\overrightarrow{Pos} \cdot \vec{i}_{z}
\end{equation}
\begin{equation}
\Delta{magnitude_{xy}}=|| \Delta\overrightarrow{Pos}_{xy} ||
\end{equation}
\begin{equation}
\Delta\theta=
\arccos{
    \frac{\Delta\overrightarrow{Pos}_{xy} \cdot \vec{i}_{x}}
    {|| \Delta\overrightarrow{Pos}_{xy} ||}    
}
\end{equation}
\begin{equation}
\Delta z=||\Delta\overrightarrow{Pos}\cdot\vec{i}_{z}||
\end{equation}

Where $\Delta\overrightarrow{Pos}_{xy}$ represents the projection of $\Delta\overrightarrow{Pos}$ on $xOy$ plane, while $\vec{i}_{z}$ and $ \vec{i}_{x}$ is the unit vector along the $z$-axis and $x$-axis. 
The horizontal movement on the $xOy$ plane is quantified by $\Delta{magnitude_{xy}}$, which is the magnitude of $\Delta\overrightarrow{Pos}_{xy}$.
The angular deviation of the camera on the plane is described by the angle $\Delta\theta$, which is the angle between $\Delta\overrightarrow{Pos}_{xy}$ and $ \vec{i}_{x}$.
Finally, $\Delta{z}$ denotes the vertical change in $z$-axis direction, which is the magnitude of  $\Delta\overrightarrow{Pos}$'s projection on this axis.

In RCM, we use $\Delta{magnitude_{xy}}$, $\Delta\theta$ and $\Delta{z}$ to control the movement of camera between two adjacent positions: 
In the RCM algorithm, the parameters $\Delta{magnitude_{xy}}$, $\Delta\theta$, and $\Delta{z}$ are utilized to navigate the camera's movement from one point to the next:

\begin{equation}
    \Delta{magnitude_{xy}}=U(a_{m},b_{m})
\end{equation}
\begin{equation}
    \Delta\theta=U(a_{\theta},b_{\theta})
\end{equation}
\begin{equation}
    \Delta{z}=U(a_{z},b_{z})
\end{equation}

In this context, $U(a_{m},b_{m})$ represents a uniform distribution utilized to generate random values for camera movement parameters. The variables $a_{m}$ and $b_{m}$ indicate the minimum and maximum limits for the horizontal movement distance on the $xOy$ plane ($\Delta{magnitude_{xy}}$), $a_{\theta}$ and $b_{\theta}$ set the limits for the angle change ($\Delta\theta$), and $a_{z}$ and $b_{z}$ are the limits for the vertical movement distance along the $z$-axis ($\Delta{z}$). These parameters ensure that the camera's position and orientation vary randomly within controlled ranges.

With each shift in camera position within the same animation scene, the parameters $\Delta{magnitude_{xy}}$, $\Delta\theta$ and $\Delta{z}$ are randomized. When the scene changes, the camera's initial position is reset to align with the new position of the character.

\begin{figure}
\begin{center}
\centering
\includegraphics[width=\linewidth]{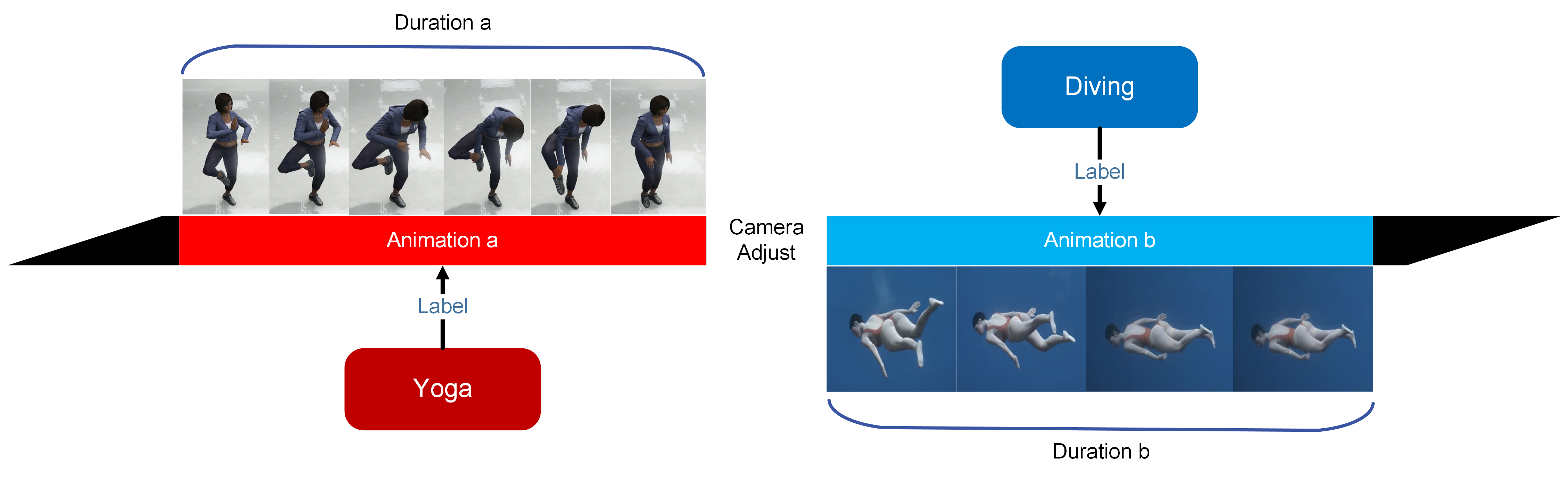}
\end{center}
\caption{Example of video clip trimming and annotation process utilizing hierarchical animation recording method.}
\vspace{-0.3cm}
\label{annotaion}
\end{figure}

\subsubsection{Hierarchical animation recording}

The Hierarchical animation recording\footnotemark[1] method organizes the captured animation videos by categories, actions, specific animations, and camera viewpoints, simplifying the video trimming and annotation process. 
Each video clip is dedicated to a single type of action, ensuring that the annotations for the target frames directly match the duration of each video clip. 
Figure~\ref{annotaion} illustrates the streamlined process of trimming and annotating using this hierarchical approach.

\section{Experimental Results}
\label{sec:results}
To assess the performance of human motion representation by GTAutoAct, we conduct a series of comparative experiments\footnotemark[1]. 
This involves evaluating the performance of models trained on datasets produced by GTAutoAct against those trained on corresponding real-world datasets in human action recognition tasks. 
Our examination encompasses representations of human motion both in major-part and whole-body.
\subsection{Human Major-part Motion Representation}
\begin{figure}
\begin{center}
\centering
\includegraphics[width=\linewidth]{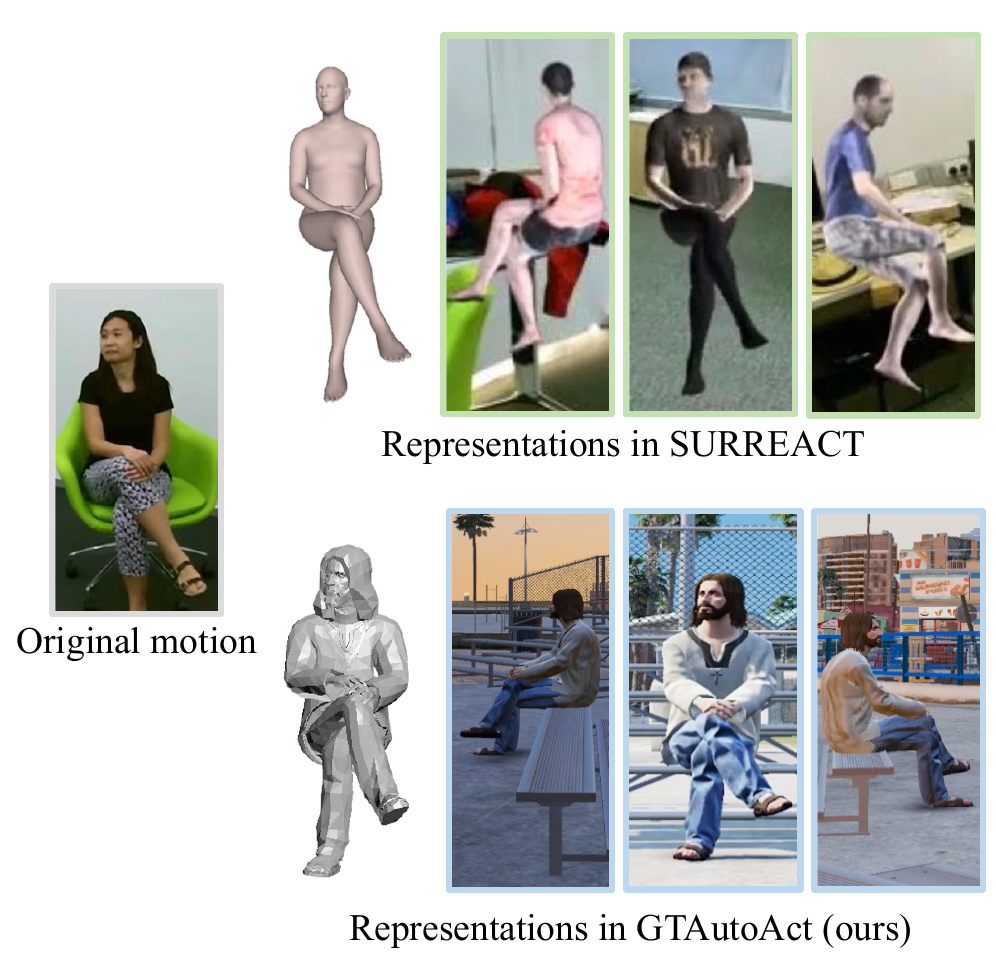}
\end{center}
\vspace{-0.5cm}
\caption{Comparison of motion representations in NTU RGB-D dataset by SURREACT and GTAutoAct.}
\label{comparison}
\vspace{-0.2cm}
\end{figure}

To evaluate the Human Major-part Motion Representation, which concentrates on the essential bone joints (Figure~\ref{configuration}) excluding the hands and feet to represent the general movements of the human body, we utilize the NTU RGB+D~\cite{ntu} action recognition dataset. 
Figure~\ref{comparison} illustrates the comparison of the action representations in NTU dataset rendered by SURREACT~\cite{synthetic_humans} and our method.

For benchmark purposes, we create \textbf{NTU-Original}, a baseline training dataset that encompasses all 49 single-person action classes from NTU RGB+D.
Additionally, we produce \textbf{NTU-GTAutoAct}, a training dataset derived from 3D human skeleton coordinates corresponding to the videos in NTU-Original utilizing GTAutoAct. 
This dataset is constructed by extracting one frame out of each five frames from each action video.
The benchmark results for various action recognition methods, evaluated by a subset of the NTU dataset \textbf{NTU-Test}, are presented in Table~\ref{ntu}.

\begin{table}
\begin{center}
\resizebox{0.5\textwidth}{!}{
\begin{tabular}{c| c c c| c c c}
\toprule[2pt]
\multicolumn{1}{c}{\bf{Model Info}} & \multicolumn{3}{c}{\bf{NTU-Original}}
& \multicolumn{3}{c}{\bf{NTU-GTAutoAct (ours)}}\\
\toprule[2pt]
\bf{Method}&\bf{Top-1}&\bf{Top-5}&\bf{Mean}&\bf{Top-1}&\bf{Top-5}&\bf{Mean}\\
\toprule[1pt]
Random &-&-&2.0 &-&-&2.0 \\
\toprule[0.6pt]
VideoMAE~\cite{videomae} &82.7&86.1&85.7 &92.6&96.4&89.3 \\
TANet~\cite{model11_tanet} &81.6&79.5&87.4 &87.6&86.2&84.6 \\
UniFormer~\cite{uniformer} &86.0&78.9&90.6 &92.4&99.7&93.2\\
TIN~\cite{model09_tin} &89.8&90.8&90.0 &89.5&95.9&95.2\\
TSN~\cite{model01_TSN} &89.1&86.6&92.0 &94.6&90.2&96.8\\
VideoSwin~\cite{videoswin} &87.9&80.3&89.2& 93.0&98.6&88.5\\
TPN~\cite{model10_tpn} &86.0&78.9&90.6 &90.2&99.5&90.2 \\
X3D~\cite{x3d} &78.2&85.2&80.9 &83.3&88.8&75.0\\
I3D~\cite{dataset05_Kinetics} &76.0&79.7&74.3& 80.1&96.3&86.5 \\
I3D NL~\cite{model05_NLI3D} &76.9&82.2&76.9& 80.3&95.4&85.4 \\
\bottomrule[2pt]
\end{tabular}
}
\end{center}
\vspace{-0.5cm}
\caption{Benchmark results on NTU-based datasets. 
Despite being input with only 20\% of the frames from the NTU-Original dataset, NTU-GTAutoAct, the dataset generated by our method, maintains high accuracy, comparable to, and in some aspects even surpassing that of NTU-Original, especially in Top-5 accuracy.}
\vspace{-0.2cm}
\label{ntu}
\end{table}

\begin{table*}
\begin{center}
\resizebox{\textwidth}{!}{
\begin{tabular}{c |c c c| c c c| c c c| c c c| c c c}
\toprule[2pt]
\multicolumn{1}{c}{\bf{Model Info}} & \multicolumn{3}{c}{\bf{H36M-Original}} & \multicolumn{3}{c}{\bf{H36M-Single}} & \multicolumn{3}{c}{\bf{H36M-Extracted}}
& \multicolumn{3}{c}{\bf{H36M-SingleExtracted}} & \multicolumn{3}{c}{\bf{H36M-GTAutoAct (Ours)}} \\
\toprule[2pt]
\bf{Method}&\bf{Top-1}&\bf{Top-5}&\bf{Mean}&\bf{Top-1}&\bf{Top-5}&\bf{Mean}&\bf{Top-1}&\bf{Top-5}&\bf{Mean}&\bf{Top-1}&\bf{Top-5}&\bf{Mean}&\bf{Top-1}&\bf{Top-5}&\bf{Mean}\\
\toprule[1pt]
Random &-&-&6.7 &-&-&6.7 &-&-&6.7 &-&-&6.7 &-&-&6.7 \\
VideoMAE~\cite{videomae} &40.2&79.2&37.5 &24.7&77.8&22.5 &15.3&50.7&17.2 &11.4&45.9&12.5 &54.5&98.2&55.0 \\
TANet~\cite{model11_tanet} &33.3&75.0&34.4 &25.5&77.5&19.7 &11.6&49.5&12.4 &13.9&50.1&14.0 &53.9&93.4&56.6 \\
UniFormer~\cite{uniformer} &37.6&76.8&38.9 &27.6&72.3&27.6 &13.7&50.4&13.7 &8.6&49.9&8.6 &52.0&92.1&53.8 \\
TIN~\cite{model09_tin} &33.8&70.1&34.2 &23.7&75.6&25.2 &13.4&47.0&11.1 &7.2&50.2&11.1 &55.5&96.3&55.1 \\
TSN~\cite{model01_TSN} &36.4&76.2&37.5 &29.5&76.8&26.4 &13.2&50.1&16.8 &12.4&56.7&12.5 &53.6&94.2&54.9 \\
VideoSwin~\cite{videoswin} &35.3&80.3&34.6 &23.6&71.1&24.6 &13.5&46.2&14.5 &14.9&50.5&12.3 &51.2&95.3&49.7 \\
TPN~\cite{model10_tpn} &35.8&78.1&33.6 &22.8&69.4&20.1 &11.4&55.3&11.4 &8.9&48.3&8.9 &61.4&96.7&55.4 \\
X3D~\cite{x3d} &32.1&67.7&29.6 &27.9&75.7&20.3 &11.2&47.3&13.2 &13.3&46.5&12.6 &47.3&90.2&46.5 \\
I3D~\cite{dataset05_Kinetics} &28.7&70.6&30.6 &20.4&69.5&19.5 &11.7&50.0&12.4 &10.5&58.5&9.6 &42.5&85.6&40.6 \\
I3D NL~\cite{model05_NLI3D} &34.2&80.5&32.4 &23.3&68.6&19.6 &11.9&50.5&11.9 &9.8&60.1&11.1 &50.6&88.9&49.7 \\
\bottomrule[2pt]
\end{tabular}
}
\end{center}
\vspace{-0.5cm}
\caption{Benchmark results on \textbf{H36M-Original-Test}. 
In addition to a notable improvement over H36M-SingleExtracted, the original dataset used for generating H36M-GTAutoAct, our method enables H36M-GTAutoAct to achieve results that surpass even those of H36M-Original, considering H36M-GTAutoAct uses only 6\% of the number of frames present in H36M-Original as input.}
\label{tabel:H36M-Segment-Test}
\vspace{0.1cm}
\end{table*}

\begin{table*}
\begin{center}
\resizebox{\textwidth}{!}{
\begin{tabular}{c| c c c| c c c| c c c| c c c| c c c}
\toprule[2pt]
\multicolumn{1}{c}{\bf{Model Info}} & \multicolumn{3}{c}{\bf{H36M-Original}} & \multicolumn{3}{c}{\bf{H36M-Single}} & \multicolumn{3}{c}{\bf{H36M-Extracted}}
& \multicolumn{3}{c}{\bf{H36M-SingleExtracted}} & \multicolumn{3}{c}{\bf{H36M-GTAutoAct (Ours)}} \\
\toprule[2pt]
\bf{Method}&\bf{Top-1}&\bf{Top-5}&\bf{Mean}&\bf{Top-1}&\bf{Top-5}&\bf{Mean}&\bf{Top-1}&\bf{Top-5}&\bf{Mean}&\bf{Top-1}&\bf{Top-5}&\bf{Mean}&\bf{Top-1}&\bf{Top-5}&\bf{Mean}\\
\toprule[1pt]
Random &-&-&6.7 &-&-&6.7 &-&-&6.7 &-&-&6.7 &-&-&6.7 \\
\toprule[0.6pt]
VideoMAE~\cite{videomae} &38.3 & 74.3 & 35.0 & 26.1 & 84.3 & 24.5 & 15.9 & 50.9 & 17.4 & 11.2 & 46.7 & 11.7 & 66.1 & 99.7 & 68.2 \\
TANet~\cite{model11_tanet} &30.9 & 81.6 & 35.6 & 24.7 & 70.7 & 18.8 & 11.7 & 52.7 & 13.6 & 14.9 & 45.2 & 14.4 & 61.7 & 85.0 & 61.9 \\
UniFormer~\cite{uniformer} &35.0 & 83.4 & 34.0 & 21.4 & 71.0 & 18.6 & 12.8 & 54.4 & 12.2 & 9.9 & 62.7 & 11.6 & 57.0 & 91.8 &57.0 \\
TIN~\cite{model09_tin} &37.1 & 75.6 & 34.4 & 25.0 & 69.4 & 23.7 & 12.3 & 49.1 & 13.8 & 15.9 & 53.5 & 13.2 & 46.7 & 88.6 & 51.1 \\
TSN~\cite{model01_TSN} &41.8 & 84.3 & 39.9 & 23.5 & 73.1 & 21.1 & 16.6 & 51.7 & 16.3 & 11.1 & 49.7 & 13.2 & 49.9 & 99.8 & 60.5 \\
VideoSwin~\cite{videoswin} &34.5 & 79.4 & 42.7 & 28.6 & 72.7 & 26.9 & 13.2 & 51.8 & 14.3 & 9.3 & 48.7 & 8.8 & 65.9 & 89.7 & 59.5 \\
TPN~\cite{model10_tpn} &29.2&80.1&33.6 &16.6&65.4&18.4 &10.6&56.7&11.6 &10.2&51.4&10.4 &75.6&98.1&68.9 \\
X3D~\cite{x3d} &30.5 & 66.9 & 34.1 & 25.3 & 69.2 & 26.0 & 13.4 & 51.6 & 10.4 & 7.5 & 54.5 & 11.3 & 58.6 & 90.3 & 50.0 \\
I3D~\cite{dataset05_Kinetics} &34.2 & 66.4 & 28.9 & 28.4 & 80.7 & 21.0 & 10.1 & 44.4 & 13.2 & 12.1 & 50.8 & 11.6 & 58.8 & 84.0 & 52.4 \\
I3D NL~\cite{model05_NLI3D} &26.4 & 72.2 & 29.3 & 19.2 & 68.0 & 20.9 & 10.8 & 51.1 & 11.7 & 9.7 & 62.4 & 9.7 & 46.6 & 87.2 & 49.5 \\
\bottomrule[2pt]
\end{tabular}
}
\end{center}
\vspace{-0.5cm}
\caption{Benchmark results on \textbf{H36M-Segment-Test}. 
Apart from the similar improvement observed in testing on H36M-Original-Test, our method enables H36M-GTAutoAct to attain a better results on segmented video data. This enhancement could be attributed to the automatic segmentation inherent in the Dynamic skeletal interpolation process.}
\vspace{-0.4cm}
\label{table:H36M-Original-Test}
\end{table*}

\subsection{Human Whole-body Motion Representation}
Human Whole-body Motion Representation by GTAutoAct includes the comprehensive modeling of all 53 bone joints as depicted in Figure~\ref{configuration}. 
To capture the 3D coordinates of human skeleton joints, we employ HRNet~\cite{hrnet}, initially trained on the H3WB~\cite{h3wb} dataset for 2D human whole-body pose estimation, which is followed by leveraging JointFormer~\cite{jointformer} for 2D-to-3D pose lifting. 

For benchmark purposes, We utilize the Human3.6M~\cite{h36m} (H36M) for comparison in whole-body motion representation. From H36M, we derive four baseline datasets of action recognition task training, as follows:
\begin{enumerate}
\item \textbf{H36M-Original}: This dataset consists of complete action videos from ``S1'', ``S5'', ``S6'', ``S7'' sessions of the H36M dataset, providing four distinct views for each action.
\item \textbf{H36M-Single}: A subset of H36M-Original, while only single view for each action is randomly selected.
\item \textbf{H36M-Extracted}: From H36M-Original, this dataset is formed by extracting one frame out of each five frames from each action video, and further randomly selecting 30\% of extracted frames, while assuring no isolated single frames, to create a sparser dataset.
\item \textbf{H36M-SingleExtracted}: Applying the same frame extraction process as H36M-Extracted, but starting from the H36M-Single dataset, to create a version that is both single-view and sparsely sampled.
\end{enumerate}
Additionally, for the purpose of testing, two specific datasets derived from different sections of the H36M dataset were used:
\begin{enumerate}
\item \textbf{H36M-Original-Test}: This test dataset includes all videos from sections ``S8'', ``S9'', ``S11'' of H36M, featuring each action captured from four distinct views.
\item \textbf{H36M-Segment-Test}:  In this dataset, every video from H36M-Original-Test has been manually segmented into individual unit actions, with the removal of indistinguishable clips.
\end{enumerate}

For our experimental purposes, we develop the \textbf{H36M-GTAutoAct} dataset for training. This dataset is created from the 3D pose estimation inferences derived from each frame of the H36M-SingleExtracted dataset, utilizing the GTAutoAct framework for generating whole-body representations. The primary objective is to assess GTAutoAct's capabilities of multi-viewpoint data generation and frame interpolation.

The benchmark results on various action recognition methods, evaluated using H36M-Original-Test and H36M-Segment-Test as test datasets, are documented in Table~\ref{table:H36M-Original-Test} and Table~\ref{tabel:H36M-Segment-Test} respectively.

\section{Conclusion}
\label{sec:conclusion}
In this paper, we introduce GTAutoAct, an innovative dataset generation framework leveraging game engine technology to facilitate advancements in action recognition. 
GTAutoAct excels in automatically creating large-scale, well-annotated datasets with extensive action classes and superior video quality in multiple viewpoints. 

Our primary contributions within GTAutoAct include:
(1) An innovative transformation from readily available coordinate-based 3D human motion into rotation-orientated whole-body representation, with enhanced suitability for diverse characters, more intuitive manipulation and realistic constraints, and simplifies interpolation.
(2) An algorithm that dynamically segments the rotation sequence into unit motions, automatically interpolates the rotation sequence across different frame counts, and randomly generates a series of variants for each animation, to ensure smooth and natural-looking animation with high diversity.
(3) An autonomous video capture and processing pipeline, featuring a randomly navigating camera, auto-trimming and labeling functionalities, all within fully customizable scenes. 

The comparative experiments we conduct, to assess the performance of human motion representation by GTAutoAct, demonstrate the significant capabilities of multi-viewpoint data generation and frame interpolation, especially under a limited number of frames as input, showing the potential for significant improvements in action recognition.

{
    \small
    \bibliographystyle{ieeenat_fullname}
    \bibliography{main}

\begin{thebibliography}{37}
\providecommand{\natexlab}[1]{#1}
\providecommand{\url}[1]{\texttt{#1}}
\expandafter\ifx\csname urlstyle\endcsname\relax
  \providecommand{\doi}[1]{doi: #1}\else
  \providecommand{\doi}{doi: \begingroup \urlstyle{rm}\Url}\fi

\bibitem[{2008-2022 OpenIV}()]{openiv}
{2008-2022 OpenIV}.
\newblock Openiv.
\newblock \url{https://openiv.com}.
\newblock Accessed March 8, 2023.

\bibitem[Ardianto and Hang(2019)]{gamedataset04_GTA360}
Sandy Ardianto and Hsueh-Ming Hang.
\newblock Nctu-gtav360: A 360° action recognition video dataset.
\newblock In \emph{2019 IEEE 21st International Workshop on Multimedia Signal Processing (MMSP)}, pages 1--5, 2019.

\bibitem[{Autodesk}()]{3dsmax}
{Autodesk}.
\newblock 3ds max 2023.
\newblock \url{https://www.autodesk.co.jp/products/3ds-max/overview?term=1-YEAR&tab=subscription}.
\newblock Accessed March 8, 2023.

\bibitem[Bloom et~al.(2012)Bloom, Makris, and Argyriou]{gamedataset01_G3D}
Victoria Bloom, Dimitrios Makris, and Vasileios Argyriou.
\newblock G3d: A gaming action dataset and real time action recognition evaluation framework.
\newblock In \emph{2012 IEEE Computer society conference on computer vision and pattern recognition workshops}, pages 7--12. IEEE, 2012.

\bibitem[Cao et~al.(2020)Cao, Gao, Mangalam, Cai, Vo, and Malik]{gamedataset02_GTAIM}
Zhe Cao, Hang Gao, Karttikeya Mangalam, Qi{-}Zhi Cai, Minh Vo, and Jitendra Malik.
\newblock Long-term human motion prediction with scene context.
\newblock \emph{CoRR}, abs/2007.03672, 2020.

\bibitem[Carreira and Zisserman(2017)]{dataset05_Kinetics}
Jo{\~{a}}o Carreira and Andrew Zisserman.
\newblock Quo vadis, action recognition? {A} new model and the kinetics dataset.
\newblock \emph{CoRR}, abs/1705.07750, 2017.

\bibitem[Carreira et~al.(2019)Carreira, Noland, Hillier, and Zisserman]{dataset15_kin700}
Jo{\~{a}}o Carreira, Eric Noland, Chloe Hillier, and Andrew Zisserman.
\newblock A short note on the kinetics-700 human action dataset.
\newblock \emph{CoRR}, abs/1907.06987, 2019.

\bibitem[{dexyfex}()]{codewroker}
{dexyfex}.
\newblock Codeworker.
\newblock \url{https://de.gta5-mods.com/tools/codewalker-gtav-interactive-3d-map}.
\newblock Accessed March 8, 2023.

\bibitem[Diba et~al.(2019)Diba, Fayyaz, Sharma, Paluri, Gall, Stiefelhagen, and Gool]{dataset01_HVU}
Ali Diba, Mohsen Fayyaz, Vivek Sharma, Manohar Paluri, J{\"{u}}rgen Gall, Rainer Stiefelhagen, and Luc~Van Gool.
\newblock Holistic large scale video understanding.
\newblock \emph{CoRR}, abs/1904.11451, 2019.

\bibitem[Fabbri et~al.(2018)Fabbri, Lanzi, Calderara, Palazzi, Vezzani, and Cucchiara]{jta}
Matteo Fabbri, Fabio Lanzi, Simone Calderara, Andrea Palazzi, Roberto Vezzani, and Rita Cucchiara.
\newblock Learning to detect and track visible and occluded body joints in a virtual world.
\newblock In \emph{European Conference on Computer Vision (ECCV)}, 2018.

\bibitem[Feichtenhofer(2020)]{x3d}
Christoph Feichtenhofer.
\newblock X3d: Expanding architectures for efficient video recognition, 2020.

\bibitem[{Fivem}()]{fivem}
{Fivem}.
\newblock Fivem.
\newblock \url{https://fivem.net}.
\newblock Accessed March 8, 2023.

\bibitem[Friedman(1984)]{supersmoother}
Jerome~H Friedman.
\newblock \emph{A variable span smoother}.
\newblock Laboratory for Computational Statistics, Department of Statistics, Stanford~…, 1984.

\bibitem[Heilbron et~al.(2015)Heilbron, Escorcia, Ghanem, and Niebles]{dataset04_ActivityNet}
Fabian~Caba Heilbron, Victor Escorcia, Bernard Ghanem, and Juan~Carlos Niebles.
\newblock Activitynet: A large-scale video benchmark for human activity understanding.
\newblock \emph{2015 IEEE Conference on Computer Vision and Pattern Recognition (CVPR)}, pages 961--970, 2015.

\bibitem[Hwang et~al.(2023)Hwang, Jang, Park, Cho, and Kim]{ElderSim}
Hochul Hwang, Cheongjae Jang, Geonwoo Park, Junghyun Cho, and Ig-Jae Kim.
\newblock Eldersim: A synthetic data generation platform for human action recognition in eldercare applications.
\newblock \emph{IEEE Access}, 11:\penalty0 9279--9294, 2023.

\bibitem[Ionescu et~al.(2014)Ionescu, Papava, Olaru, and Sminchisescu]{h36m}
Catalin Ionescu, Dragos Papava, Vlad Olaru, and Cristian Sminchisescu.
\newblock Human3.6m: Large scale datasets and predictive methods for 3d human sensing in natural environments.
\newblock \emph{IEEE Transactions on Pattern Analysis and Machine Intelligence}, 2014.

\bibitem[Kuehne et~al.(2011)Kuehne, Jhuang, Garrote, Poggio, and Serre]{dataset02_HMDB}
H. Kuehne, H. Jhuang, E. Garrote, T. Poggio, and T. Serre.
\newblock Hmdb: A large video database for human motion recognition.
\newblock In \emph{2011 International Conference on Computer Vision}, pages 2556--2563, 2011.

\bibitem[Li et~al.(2022)Li, Wang, Peng, Song, Liu, Li, and Qiao]{uniformer}
Kunchang Li, Yali Wang, Gao Peng, Guanglu Song, Yu Liu, Hongsheng Li, and Yu Qiao.
\newblock Uniformer: Unified transformer for efficient spatial-temporal representation learning.
\newblock In \emph{International Conference on Learning Representations}, 2022.

\bibitem[Lin et~al.(2019)Lin, Gan, and Han]{model02_TSM}
Ji Lin, Chuang Gan, and Song Han.
\newblock Tsm: Temporal shift module for efficient video understanding.
\newblock In \emph{Proceedings of the IEEE International Conference on Computer Vision}, 2019.

\bibitem[Liu et~al.(2020)Liu, Wang, Wu, Qian, and Lu]{model11_tanet}
Zhaoyang Liu, Limin Wang, Wayne Wu, Chen Qian, and Tong Lu.
\newblock Tam: Temporal adaptive module for video recognition.
\newblock \emph{arXiv preprint arXiv:2005.06803}, 2020.

\bibitem[Liu et~al.(2022)Liu, Ning, Cao, Wei, Zhang, Lin, and Hu]{videoswin}
Ze Liu, Jia Ning, Yue Cao, Yixuan Wei, Zheng Zhang, Stephen Lin, and Han Hu.
\newblock Video swin transformer.
\newblock In \emph{Proceedings of the IEEE/CVF Conference on Computer Vision and Pattern Recognition}, pages 3202--3211, 2022.

\bibitem[Long et~al.(2020)Long, Yao, Qiu, Tian, Luo, and Mei]{dataset14_kin600}
Fuchen Long, Ting Yao, Zhaofan Qiu, Xinmei Tian, Jiebo Luo, and Tao Mei.
\newblock Learning to localize actions from moments.
\newblock \emph{CoRR}, abs/2008.13705, 2020.

\bibitem[Lutz et~al.(2022)Lutz, Blythman, Ghostal, Matthew, Simms, and Smolic]{jointformer}
Sebastian Lutz, Richard Blythman, Koustav Ghostal, Moynihan Matthew, Ciaran Simms, and Aljosa Smolic.
\newblock Jointformer: Single-frame lifting transformer with error prediction and refinement for 3d human pose estimation.
\newblock \emph{26TH International Conference on Pattern Recognition, {ICPR} 2022}, 2022.

\bibitem[Monfort et~al.(2019)Monfort, Andonian, Zhou, Ramakrishnan, Bargal, Yan, Brown, Fan, Gutfruend, Vondrick, et~al.]{dataset10_mit}
Mathew Monfort, Alex Andonian, Bolei Zhou, Kandan Ramakrishnan, Sarah~Adel Bargal, Tom Yan, Lisa Brown, Quanfu Fan, Dan Gutfruend, Carl Vondrick, et~al.
\newblock Moments in time dataset: one million videos for event understanding.
\newblock \emph{IEEE Transactions on Pattern Analysis and Machine Intelligence}, pages 1--8, 2019.

\bibitem[Monfort et~al.(2021)Monfort, Pan, Ramakrishnan, Andonian, McNamara, Lascelles, Fan, Gutfreund, Feris, and Oliva]{dataset11_mmit}
Mathew Monfort, Bowen Pan, Kandan Ramakrishnan, Alex Andonian, Barry~A McNamara, Alex Lascelles, Quanfu Fan, Dan Gutfreund, Rog{\'e}rio~Schmidt Feris, and Aude Oliva.
\newblock Multi-moments in time: Learning and interpreting models for multi-action video understanding.
\newblock \emph{IEEE Transactions on Pattern Analysis and Machine Intelligence}, 44\penalty0 (12):\penalty0 9434--9445, 2021.

\bibitem[Roitberg et~al.(2021)Roitberg, Schneider, Djamal, Seibold, Rei{\ss}, and Stiefelhagen]{gamedataset03_SIM}
Alina Roitberg, David Schneider, Aulia Djamal, Constantin Seibold, Simon Rei{\ss}, and Rainer Stiefelhagen.
\newblock Let's play for action: Recognizing activities of daily living by learning from life simulation video games.
\newblock \emph{CoRR}, abs/2107.05617, 2021.

\bibitem[Shahroudy et~al.(2016)Shahroudy, Liu, Ng, and Wang]{ntu}
Amir Shahroudy, Jun Liu, Tian-Tsong Ng, and Gang Wang.
\newblock Ntu rgb+d: A large scale dataset for 3d human activity analysis.
\newblock In \emph{Proceedings of the IEEE conference on computer vision and pattern recognition}, pages 1010--1019, 2016.

\bibitem[Shao et~al.(2020{\natexlab{a}})Shao, Zhao, Dai, and Lin]{dataset13_finegym}
Dian Shao, Yue Zhao, Bo Dai, and Dahua Lin.
\newblock Finegym: A hierarchical video dataset for fine-grained action understanding.
\newblock In \emph{Proceedings of the IEEE/CVF Conference on Computer Vision and Pattern Recognition}, pages 2616--2625, 2020{\natexlab{a}}.

\bibitem[Shao et~al.(2020{\natexlab{b}})Shao, Qian, and Liu]{model09_tin}
Hao Shao, Shengju Qian, and Yu Liu.
\newblock Temporal interlacing network.
\newblock \emph{AAAI}, 2020{\natexlab{b}}.

\bibitem[Soomro et~al.(2012)Soomro, Zamir, and Shah]{dataset03_UCF}
K. Soomro, A. Zamir, and M. Shah.
\newblock Ucf101: A dataset of 101 human actions classes from videos in the wild.
\newblock \emph{ArXiv}, abs/1212.0402, 2012.

\bibitem[Sun et~al.(2019)Sun, Xiao, Liu, and Wang]{hrnet}
Ke Sun, Bin Xiao, Dong Liu, and Jingdong Wang.
\newblock Deep high-resolution representation learning for human pose estimation.
\newblock In \emph{Proceedings of the IEEE/CVF Conference on Computer Vision and Pattern Recognition (CVPR)}, 2019.

\bibitem[Tong et~al.(2022)Tong, Song, Wang, and Wang]{videomae}
Zhan Tong, Yibing Song, Jue Wang, and Limin Wang.
\newblock Video{MAE}: Masked autoencoders are data-efficient learners for self-supervised video pre-training.
\newblock In \emph{Advances in Neural Information Processing Systems}, 2022.

\bibitem[Varol et~al.(2021)Varol, Laptev, Schmid, and Zisserman]{synthetic_humans}
G{\"u}l Varol, Ivan Laptev, Cordelia Schmid, and Andrew Zisserman.
\newblock Synthetic humans for action recognition from unseen viewpoints.
\newblock In \emph{IJCV}, 2021.

\bibitem[Wang et~al.(2016)Wang, Xiong, Wang, Qiao, Lin, Tang, and Van~Gool]{model01_TSN}
Limin Wang, Yuanjun Xiong, Zhe Wang, Yu Qiao, Dahua Lin, Xiaoou Tang, and Luc Van~Gool.
\newblock Temporal segment networks: Towards good practices for deep action recognition.
\newblock In \emph{European conference on computer vision}, pages 20--36. Springer, 2016.

\bibitem[Wang et~al.(2018)Wang, Girshick, Gupta, and He]{model05_NLI3D}
Xiaolong Wang, Ross Girshick, Abhinav Gupta, and Kaiming He.
\newblock Non-local neural networks.
\newblock \emph{CVPR}, 2018.

\bibitem[Yang et~al.(2020)Yang, Xu, Shi, Dai, and Zhou]{model10_tpn}
Ceyuan Yang, Yinghao Xu, Jianping Shi, Bo Dai, and Bolei Zhou.
\newblock Temporal pyramid network for action recognition.
\newblock In \emph{Proceedings of the IEEE Conference on Computer Vision and Pattern Recognition (CVPR)}, 2020.

\bibitem[Zhu et~al.(2023)Zhu, Samet, and Picard]{h3wb}
Yue Zhu, Nermin Samet, and David Picard.
\newblock H3wb: Human3.6m 3d wholebody dataset and benchmark.
\newblock In \emph{Proceedings of the IEEE/CVF International Conference on Computer Vision (ICCV)}, pages 20166--20177, 2023.

\end{thebibliography}
}

\clearpage

\setcounter{page}{1}
\setcounter{section}{0}
\maketitleappendix
\renewcommand\thesection{\Alph{section}}
\section{Extended Introduction to GTAutoAct}
This section offers supplementary details to further elucidate the GTAutoAct introduction.

\subsection{Interpretation of Terminologies}
\begin{figure}
\begin{center}
\centering
\includegraphics[width=\linewidth]{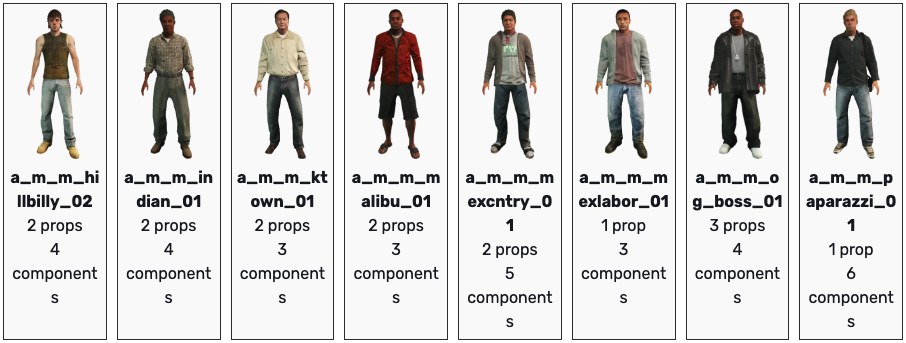}
\end{center}
\caption{Examples of native Peds.}
\label{ped}
\end{figure}

In GTAutoAct, we define several key terms to facilitate the creation and modification of action recognition datasets:
\begin{itemize}
    \item \textbf{Character}: Refers to the main player executing the action, captured by the camera.
    \item \textbf{Scene}: Denotes the environment where the action takes place. It includes natural game elements like weather and time of day, character conditions such as appearance and behavior style, and the scenery within the camera's view, such as character location and surroundings.
    \item \textbf{Animation}: Represents the sequence of movements or actions performed by the character.
    \item \textbf{Entity}: Encompasses all in-game objects, ranging from large structures like buildings to small items like books. Entities possess attributes beyond appearance, including weight, texture, collision volume, etc. 
    \item \textbf{Ped (Pedestrian)}: In GTAV, this term refers to non-player characters or pedestrian models within the game. FiveM allows players to customize their in-game pedestrian models. Examples of native Peds are illustrated in Figure~\ref{ped}.
\end{itemize}

\subsection{Theoretical Underpinnings}
This section provides additional explanations of the concepts introduced in the main body of the paper.
\subsubsection{Rotational degree of freedom}

\begin{figure}
\begin{center}
\centering
\includegraphics[width=0.7\linewidth]{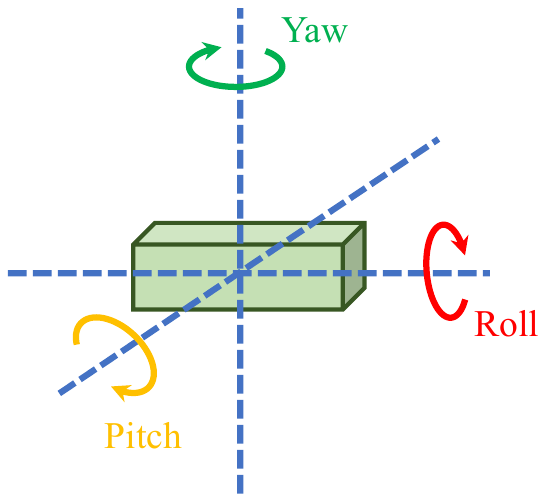}
\end{center}
\caption{Schematic diagram of three degrees of freedom of a flying drone.}
\label{freedom}
\end{figure}

A degree of freedom in general is an independent parameter that defines the state of a system. In the context of rotation, a rotational degree of freedom is a specific way in which a system or object can rotate.
In a three-dimensional space, a rigid body possesses a maximum of three rotational degrees of freedom, as it can independently rotate around three orthogonal axes, commonly identified as the $X$, $Y$, and $Z$ axes in a Cartesian coordinate system.
For instance, a standard flying drone exhibits three rotational degrees of freedom: pitch (tilting up and down), yaw (turning left or right), and roll (tilting side to side). These movements are depicted in Figure~\ref{freedom}.

\subsubsection{Rotation angle transition}
The rotation axis $\Vec{a}$ and angle $\theta$ that transform a vector from its initial position $\Vec{v}_0$ to it current position $\Vec{v}_1$ can be calculated as following:
\begin{equation}
    \Vec{a} = \Vec{v}_0 \times \Vec{v}_1
\end{equation}
\begin{equation}
    \theta = \arccos{\frac{\Vec{v}_0\cdot\Vec{v}_1}{\Vert\Vec{v}_0\Vert \cdot \Vert\Vec{v}_1\Vert}}
\end{equation}

Additionally, the rotation by an angle $\theta$, denoted as rotation matrix $RM(\theta)$, can be represented as a composition of rotations about the $x$, $y$, and $z$ axes, with respective rotation angles $\alpha$, $\beta$, and $\gamma$. This is accomplished using a rotation matrix, which combines the individual rotation matrices for each of these axes. The resulting rotation matrix, representing the combined rotation, is given by:
\begin{equation}
    RM(\theta) = RM_x(\alpha) \cdot RM_y(\beta) \cdot RM_z(\gamma)
\end{equation}

\begin{equation}
RM(\theta)=
    \begin{pmatrix}
    \cos\theta & -\sin\theta & 0 \\
    \sin\theta & \cos\theta & 0\\
    0 & 0 & 1 
    \end{pmatrix}
\end{equation}

\subsubsection{Dynamic skeletal interpolation algorithm}
In the Dynamic Skeletal Interpolation Algorithm, the Random Variation function $\mathcal{V}(\cdot)$ is used to generate a series of variants for each animation to ensure the diversity of each action, which can be denoted as following:
\begin{equation}
    \mathcal{V}(\textbf{A}, V)=\{ \mathcal{A}_v(\textbf{q}_{b}^{f}+\delta) | v\in V, b\in B, f \in F\}
\end{equation}
\begin{equation}
    \delta = U(a_{q}, b_{q})
\end{equation}
In this context, $\textbf{A}$ represents the original animation, while $V$ signifies the number of variants created.
The function $\mathcal{A}_v(\cdot)$ denotes the variant $v$ of the animation, characterized by the variant quaternion $\textbf{q}_{b}^{f}$ for each bone joint $b$ in frame $f$. 
Here, $B$ corresponds to the total number of bone joints in the animation, and $F$ indicates the total number of frames within that animation.

\subsubsection{Reference vector}
\begin{figure}
\begin{center}
\centering
\includegraphics[width=0.8\linewidth]{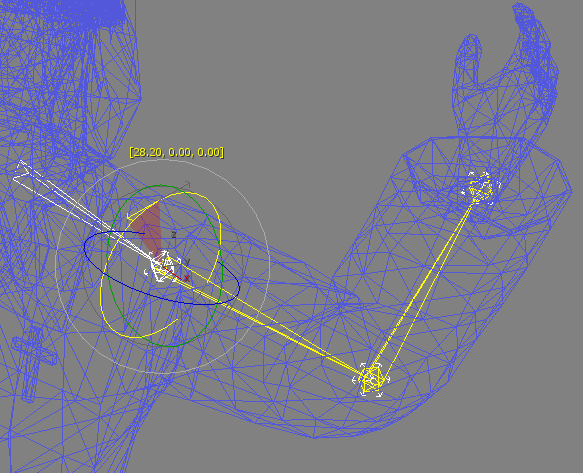}
\end{center}
\caption{A schematic diagram illustrates the role of the reference vector (representing the forearm) in calculating the rotation angle of the target vector (representing the upper arm). The rotation of the target vector around its own axis is represented as the rotation of the reference vector around the target vector. This rotation is depicted by the red circular sector in the figure.}
\label{ref_vec}
\end{figure}

Representing a three-dimensional object, characterized by its volume, shape, and surface details, solely with a vector in the same space, can lead to the loss of some of its dimensional attributes.
A vector in three-dimensional space is typically used to represent direction and magnitude but falls short of capturing the complete intricacies of a three-dimensional object's shape or volume. 
Consequently, reducing an object to a mere vector results in the omission of certain spatial characteristics and nuances that are integral to the object's full representation.
Therefore, to enhance this simplified representation, we introduce a reference vector, illustrated in Figure~\ref{ref_vec} . 
This addition is designed to incorporate or compensate for the dimensional details and aspects that the initial vector-based representation omits.
By utilizing this reference vector, we can accurately calculate the rotation of the object around itself.

\subsection{Foundational Mathematical Concepts}
This section provides essential mathematical background and supplementary details for the concepts discussed in the main paper.
\subsubsection{Euler angle}
Euler angles provide a way to express the orientation of a rigid body within three-dimensional space. 
Euler angles describe a sequence of three rotations around designated axes, which bring it into a specified orientation when executed on a coordinate system or a rigid body.

Based on the axes chosen for rotation representation and computation, Euler angles can be categorized into two distinct groups:
\begin{itemize}
    \item \textbf{Proper Euler Angles}: In the Proper Euler Angles representation, the rotations are performed about two distinct axes. Usually, the first and third rotations revolve around the same axis, whereas the second rotation involves a different axis. A typical notation for Proper Euler Angles is $Z-X-Z$, indicting that the first rotation is around the $Z$-axis, the second around the $X$-axis, and the third around the $Z$-axis. Proper Euler angles are frequently employed in situations where symmetry is crucial, such as  in the rotational movements of rigid bodies like gyroscopes and celestial objects.
    \item  \textbf{Tait-Bryan Angles}: Tait-Bryan angles encompass rotations around three distinct axes, typically with each rotation occurring about a different axis. A frequently used sequence in Tait-Bryan angles is Yaw-Pitch-Roll ($Y-X-Z$), which signifies rotations about the $Y$-axis, $X$-axis, and $Z$-axis respectively. These angles are generally more intuitive for visualizing and comprehending rotations in physical space and are commonly applied in fields such as robotics, camera movements in computer graphics, and vehicle dynamics.
\end{itemize}
Based on the specific axes used for each elemental rotation, the rotational representations expressed by Euler angles can be categorized into two distinct groups:
\begin{itemize}
    \item \textbf{Intrinsic Rotations}: In intrinsic rotations, the rotations are applied relative to the object's own coordinate system, which alters its orientation after each rotation. Consequently, each subsequent rotation occurs around an axis of the newly oriented coordinate system.
    \item \textbf{Extrinsic Rotations}: In extrinsic rotations, the rotations are performed with respect to a stationary coordinate system. As a result, each rotation occurs around an axis of this fixed (or world) coordinate system, which remains unchanged regardless of the object's rotations.
\end{itemize}

In GTAutoAct, we utilize the Tait-Bryan Angles of Extrinsic Rotations, following the $x-y-z$ sequence. These are defined as Static Euler Angles, to represent the rotation of each bone joint. 

\subsubsection{Quaternion}
A quaternion is a mathematical construct that extends the concept of complex numbers. 
It is predominantly utilized in three-dimensional computing, notably in fields like computer graphics and robotics, where it facilitates efficient management of rotations and orientations. 
Defining two quaternions:

\begin{equation}
    \textbf{q}_1=a_1\textbf{i}+b_1\textbf{j}+c_1\textbf{k}+w_1
\end{equation}
\begin{equation}
    \textbf{q}_2=a_2\textbf{i}+b_2\textbf{j}+c_2\textbf{k}+w_2
\end{equation}
their addition is expressed as a component-wise addition:
\begin{equation}
    \textbf{q}_1+\textbf{q}_2=(a_1+a_2)\textbf{i}+(b_1+b_2)\textbf{j}+(c_1+c_2)\textbf{k}+(w_1+w_2)
\end{equation}
For these two quaternions, their multiplication is represented by the Hamilton product, 
\begin{equation}
\begin{aligned}
    \textbf{q}_1\times\textbf{q}_2=w_1w_2-a_1a_2-b_1b_2-c_1c_2\\
    +(w_1a_2+a_1w_2+b_1c_2-c_1b_2)\textbf{i}\\
    +(w_1b_2-a_1c_2+b_1w_2+c_1a_2)\textbf{j}\\
    +(w_1c_2+a_1b_2-b_1a_2+c_1w_2)\textbf{k}\\
\end{aligned}
\end{equation}
The conjugation of a quaternion $\textbf{q}$ is expressed by reversing the sign of its vector part while keeping the scalar part unchanged:
\begin{equation}
    \textbf{q}^*=Conj(\textbf{q})=-\frac{1}{2}(\textbf{q}+\textbf{i}\textbf{q}\textbf{i}+\textbf{j}\textbf{q}\textbf{j}+\textbf{k}\textbf{q}\textbf{k})
\end{equation}
The multiplication inverse (or reciprocal) of a quaternion $\textbf{q}$ can be represented as follows:
\begin{equation}
    \textbf{q}^{-1}=Inv(\textbf{q})=\frac{\textbf{q}^*}{\Vert \textbf{q} \Vert^2}
\end{equation}

\subsubsection{Transformation matrix}
A transformation matrix is a matrix that facilitates the linear transformation of a geometric figure. 
Utilizing homogeneous coordinates to represent points is a common practice, as it enables more intricate transformations, such as perspective projections, to be efficiently expressed through matrix multiplications.

Therefore, the $4\times4$ transformation matrix $A$, representing transformation $T(\cdot)$, can be expressed using a combination of a $3\times3$ rotation matrix $RM$, a $3\times1$ translation matrix $TM$, and a $1\times3$ zero matrix. The structure of the matrix $A$ is typically as follows:
\begin{equation}
    T(\textbf{x}) = A\textbf{x}
\end{equation}

\begin{equation}
A=
    \begin{pmatrix}
    RM & TM \\
    \textbf{0} & 1
    \end{pmatrix}
\end{equation}

\subsection{Extended Introductions}
This section presents detailed elaborations on the modules and functionalities referenced in the main paper.
\subsubsection{Mesh generation process}
Initially, we employ OpenIV~\cite{openiv}, a popular modification tool for GTAV and FiveM servers, which enables users to access and modify a range of game files such as textures, models, and other assets. 
Using this tool, we export native pedestrian models (peds) from the built-in library.
Subsequently, we use 3ds Max~\cite{3dsmax}, a professional 3D computer graphics program renowned for creating 3D animations, models, games, and images.
With 3ds Max, we craft a variety of Meshes from the exported peds, applying different configurations to effectively display animations. 

\subsubsection{Scene customization}

\begin{figure}
\begin{center}
\centering
\includegraphics[width=\linewidth]{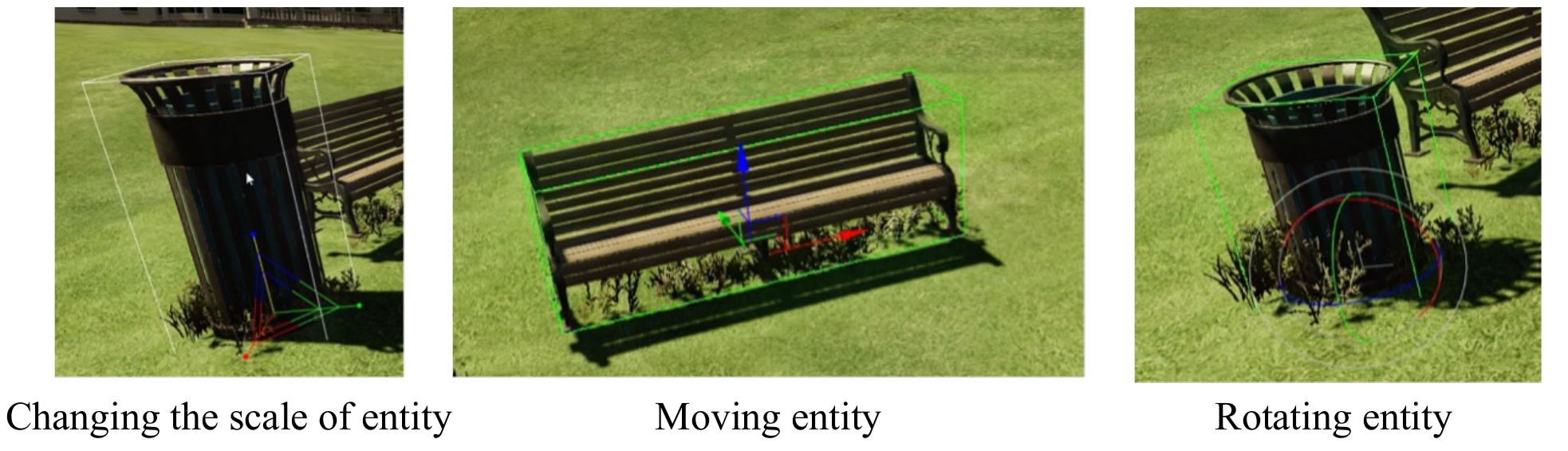}
\end{center}
\caption{Adjusting entity in CodeWorker.}
\label{codeworker}
\end{figure}

We utilize CodeWorker~\cite{codewroker}, a versatile tool designed for exploring, modifying, and exporting map data in GTAV and FiveM servers. 
This tool enables us to efficiently edit maps by adjusting various entities within the game. 
The process and capabilities of CodeWorker in facilitating these modifications are illustrated in Figure~\ref{codeworker}.
Additionally, CodeWorker offers functionality to import self-created, customized entities that have been developed using game engines.

In the aspect of Environmental Customization, we have chosen 14 different types of weather changes, as detailed in Table~\ref{tabweather}, and 4 distinct time changes, corresponding to four specific periods of the day, which are outlined in Table~\ref{tabtime}, as illustrative examples. 
Through the customization of weather and clock time settings, GTAutoAct is equipped to represent varying natural conditions within the game, encompassing aspects such as illumination and shadow.

\begin{table}
\begin{center}
\begin{tabular}{c c c}
\bf{Weather} & \bf{Example} & \bf{Weather} \\
BLIZZARD & 
    \begin{minipage}[b]{0.3\columnwidth}
		\centering
		\raisebox{-.5\height}{\includegraphics[width=\linewidth]{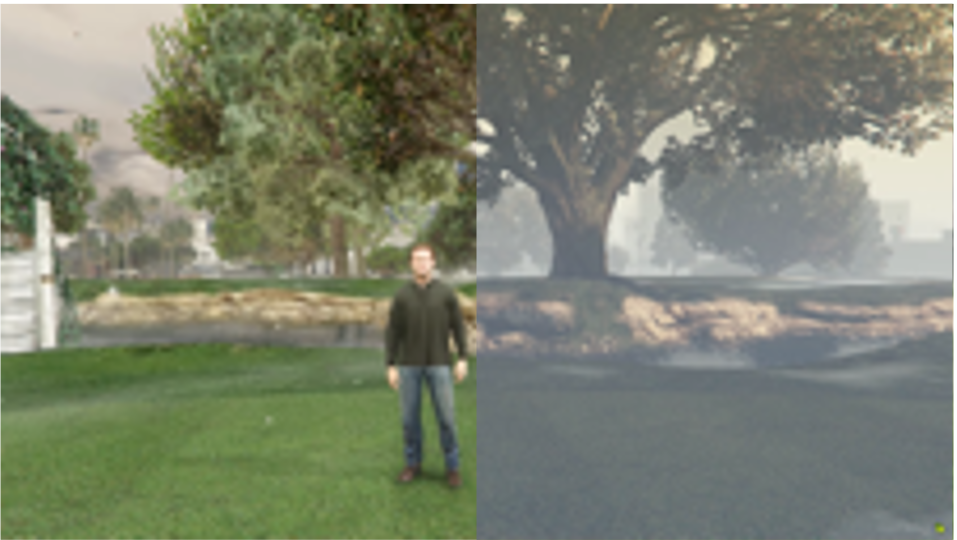}}
	\end{minipage} 
 & NEUTRAL \\
CLEAR &     
    \begin{minipage}[b]{0.3\columnwidth}
		\centering
		\raisebox{-.5\height}{\includegraphics[width=\linewidth]{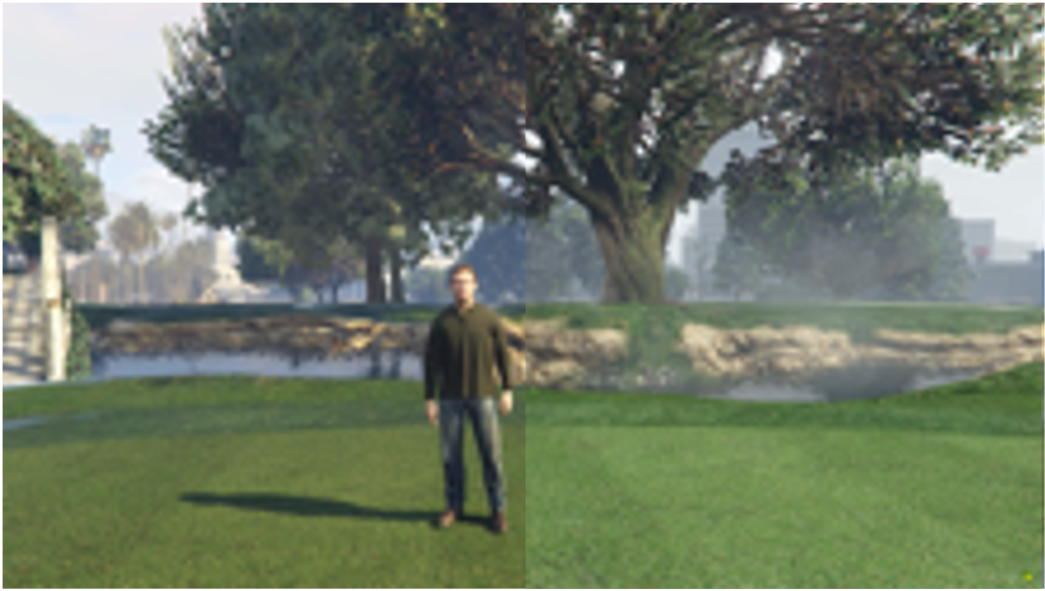}}
	\end{minipage}  
 & OVERCAST \\
 CLEARING &     
    \begin{minipage}[b]{0.3\columnwidth}
		\centering
		\raisebox{-.4\height}{\includegraphics[width=\linewidth]{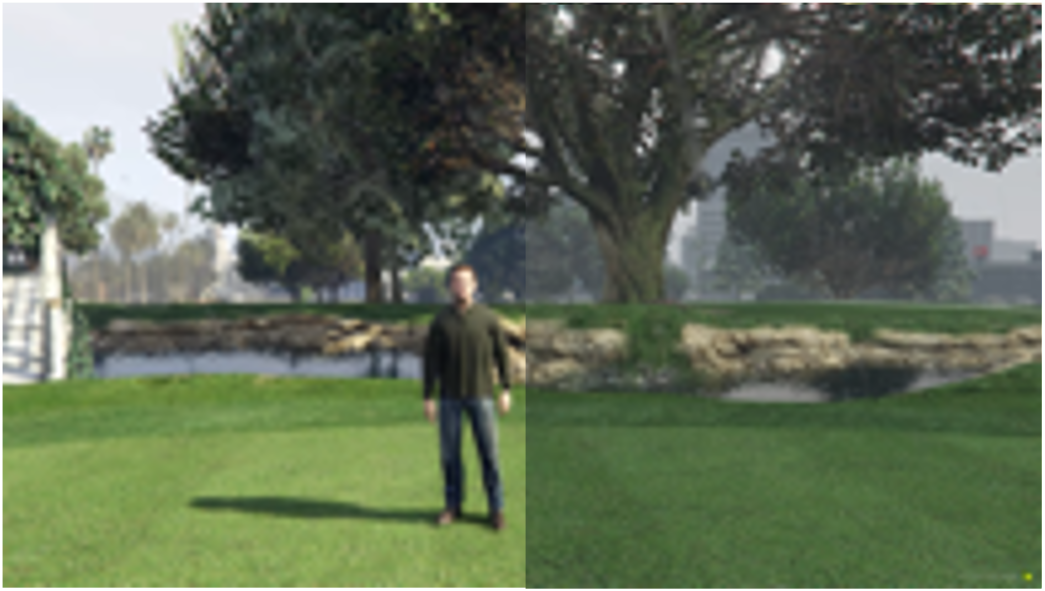}}
	\end{minipage}  
 & RAIN \\
 CLOUDS &     
    \begin{minipage}[b]{0.3\columnwidth}
		\centering
		\raisebox{-.5\height}{\includegraphics[width=\linewidth]{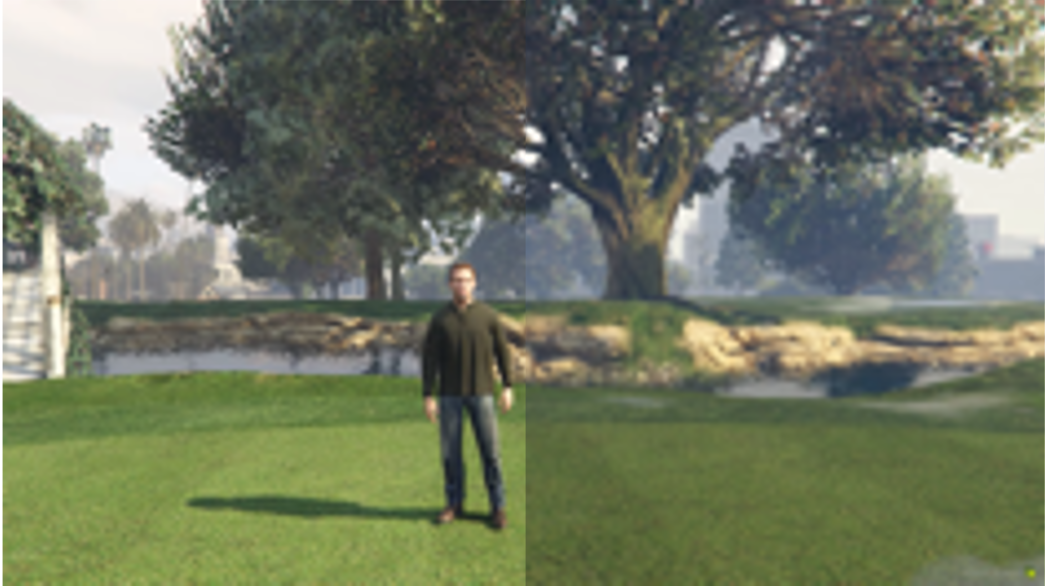}}
	\end{minipage}  
 & SMOG \\
EXTRASUNNY &     
    \begin{minipage}[b]{0.3\columnwidth}
		\centering
		\raisebox{-.5\height}{\includegraphics[width=\linewidth]{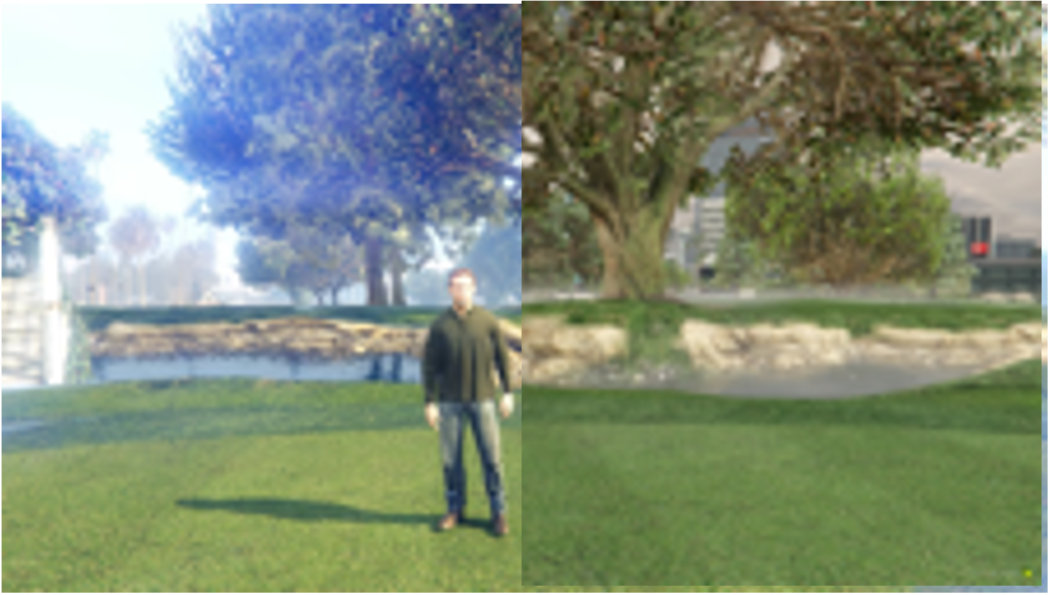}}
	\end{minipage}  
 & SNOW \\
 FOGGY &     
    \begin{minipage}[b]{0.3\columnwidth}
		\centering
		\raisebox{-.5\height}{\includegraphics[width=\linewidth]{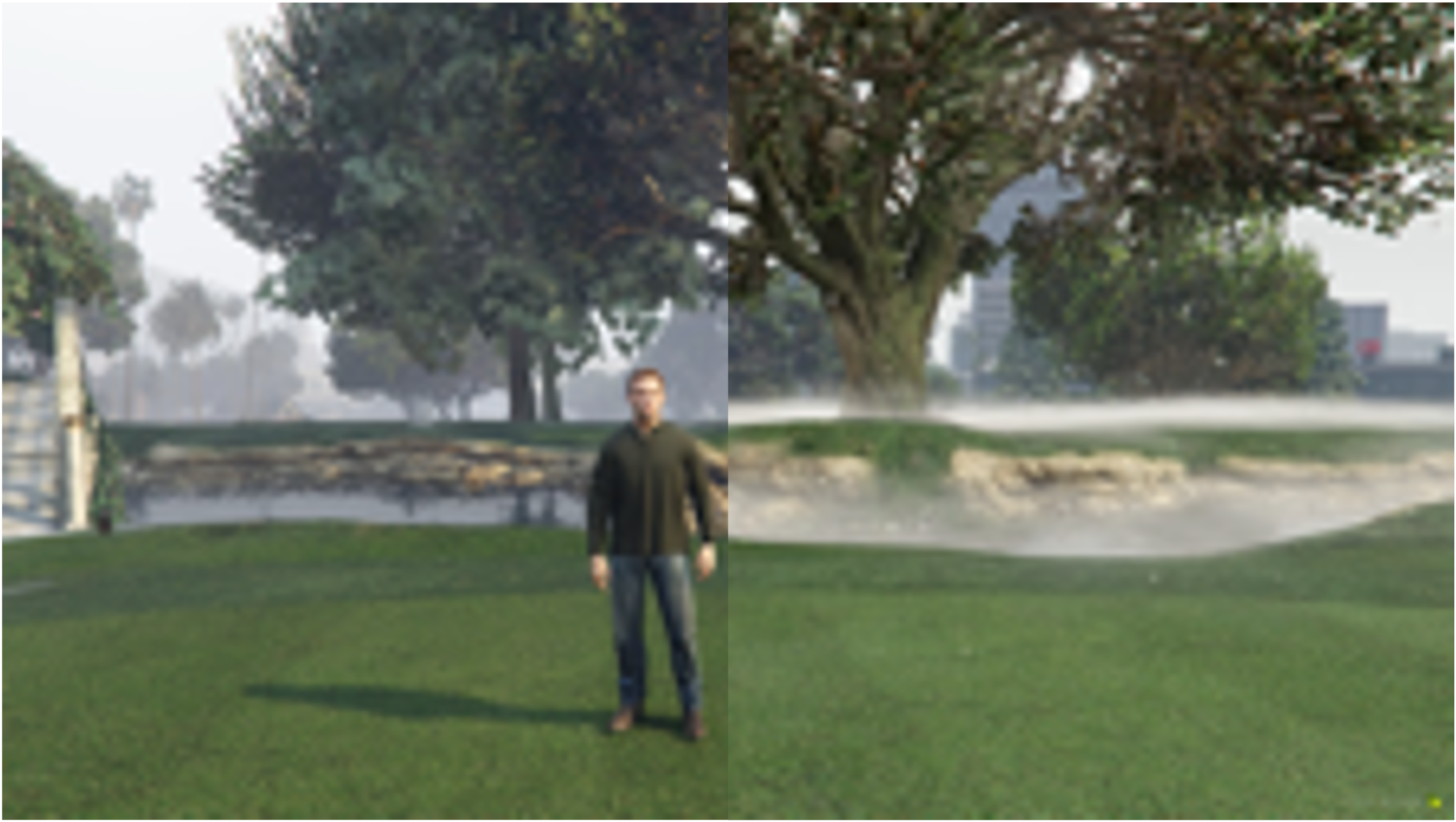}}
	\end{minipage}  
 & SNOWLIGHT \\
  XMAS &     
    \begin{minipage}[b]{0.3\columnwidth}
		\centering
		\raisebox{-.5\height}{\includegraphics[width=\linewidth]{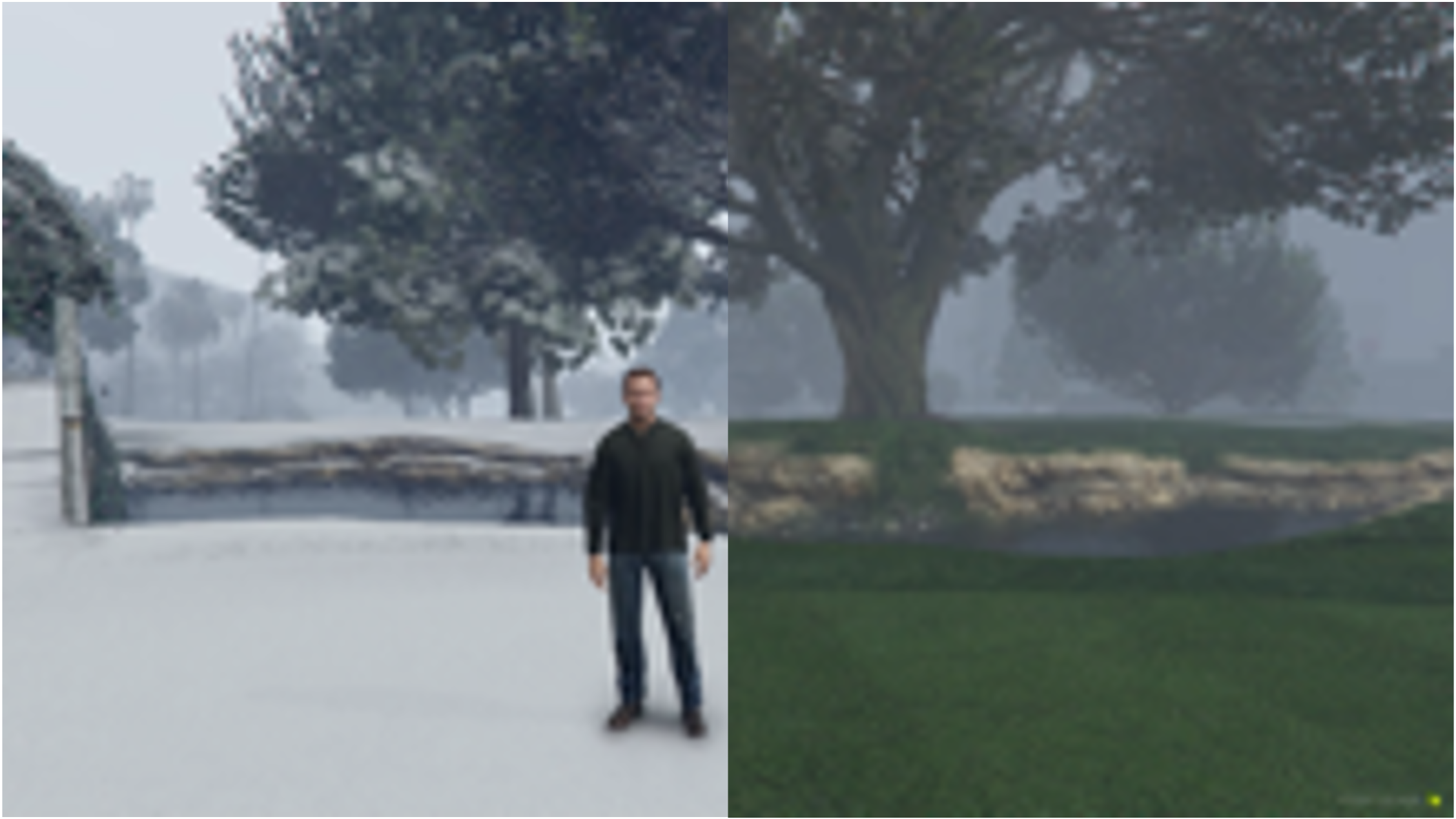}}
	\end{minipage}  
 & THUNDER \\
\end{tabular}
\end{center}
\caption{Examples of alternative weathers in GTAutoAct}
\label{tabweather}
\end{table}

\begin{table}
\begin{center}
\begin{tabular}{c c c}
\bf{Time} & \bf{Period} & \bf{Example} \\
5:30 & Dawn &
    \begin{minipage}[b]{0.3\columnwidth}
		\centering
		\raisebox{-.5\height}{\includegraphics[width=\linewidth]{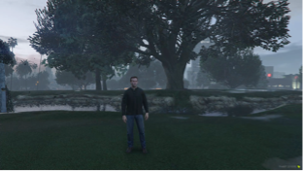}}
	\end{minipage} \\
12:00 & Noon &
    \begin{minipage}[b]{0.3\columnwidth}
		\centering
		\raisebox{-.5\height}{\includegraphics[width=\linewidth]{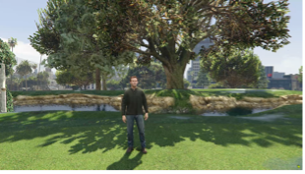}}
	\end{minipage} \\
 20:00 & Evening &     
    \begin{minipage}[b]{0.3\columnwidth}
		\centering
		\raisebox{-.4\height}{\includegraphics[width=\linewidth]{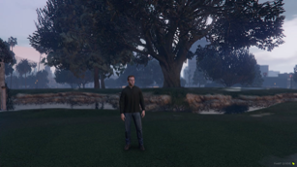}}
	\end{minipage} \\
 23:00 & Midnight &
    \begin{minipage}[b]{0.3\columnwidth}
		\centering
		\raisebox{-.5\height}{\includegraphics[width=\linewidth]{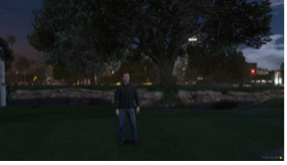}}
	\end{minipage} \\
\end{tabular}
\end{center}
\caption{Example of alternative times in GTAutoAct}
\label{tabtime}
\end{table}

\subsubsection{Hierarchical animation recording}
\begin{figure*}
\begin{center}
\includegraphics[width=1\textwidth]{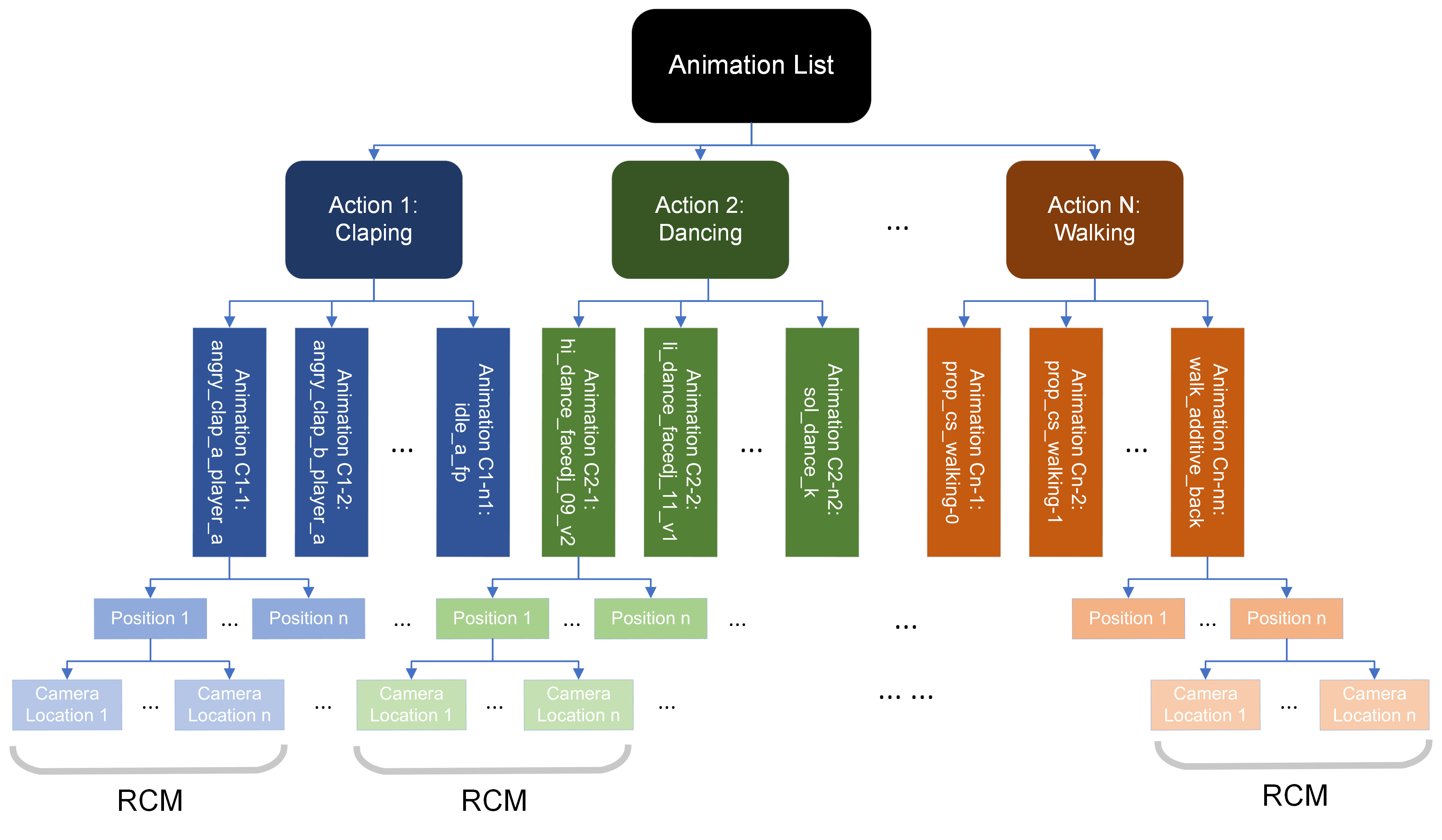}
\end{center}
   \caption{Examples of hierarchical animation recording in GTAutoAct.}
\label{HAR}
\end{figure*}
As for recording, we begin by categorizing the animation list into different classes based on the actions, and then create a sub-list for each class, as depicted in the second line of Figure~\ref{HAR}. 
Within each class, every animation in the sub-list is demonstrated in positions appropriate to the specific action, illustrated in the fourth line of Figure~\ref{HAR}. 
For instance, the action ``diving'' is showcased exclusively in underwater locations.
For each chosen position, we employs the Random Camera Moving (RCM) algorithm to automatically orchestrate camera movements, as highlighted in the fifth line of Figure~\ref{HAR}. 
Concurrently, the animation is being recorded, capturing the action from dynamically changing perspectives.

\section{Experiment Configurations}
This section provides comprehensive details about the experimental setups mentioned in the main paper.

\subsection{Experimental Environments}
This section introduces the experimental environments used for the evaluations.
\begin{itemize}
    \item \textbf{Wisteria/BDEC-01 Supercomputer System}:\\
    CPU: Intel$^\circledR$ Xeon$^\circledR$ Gold 6258R CPU @2.70GHz;\\
    CPU number: 112;\\
    Memory: 512 GiB;\\
    GPU: NVIDIA A100-SXM4-40GB;\\
    GPU number: 8;\\
    System: Red Hat Enterprise Linux 8;\\
    Python: 3.8.18; \\
    NVCC: V11.6.124;\\
    GCC: 8.3.1 20191121 (Red Hat 8.3.1-5);\\
    PyTorch: 1.13.1;\\
    CUDA: 11.6;\\
    CuDNN: 8.3.2.\\ 

    \item \textbf{Server}:\\
    CPU: Intel$^\circledR$ Xeon$^\circledR$ W-2125 CPU @4.00GHz;\\
    CPU number: 8;\\
    Memory: 128 GB;\\
    GPU: Quadro RTX 6000;\\
    GPU number: 1;\\
    System: Ubuntu 20.04.6 LTS;\\
    Python: 3.7.16; \\
    NVCC: V11.3.58;\\
    GCC: (Ubuntu 9.4.0-1ubuntu1~20.04.2) 9.4.0;\\
    PyTorch: 1.11.0+cu113;\\
    CUDA: 11.3;\\
    CuDNN: 8.2.\\ 

    \item \textbf{PC}:\\
    CPU: Intel63 Family 6 Model 158 Stepping 9 GenuineIntel @3.6GHz;\\
    CPU number: 1;\\
    Memory: 32 GB;\\
    GPU: NVIDIA GeForce RTX 3060;\\
    GPU number: 1;\\
    System: Microsoft Windows 10.0.19045;\\
    Python: 3.8.17; \\
    NVCC: V11.8.89;\\
    GCC: N/A;\\
    PyTorch: 2.0.0;\\
    CUDA: 11.8;\\
    CuDNN: 8.7.\\ 

\end{itemize}

\subsection{Implementation Details}
This section details the implementation specifics for the various tasks outlined in the main paper.
\begin{itemize}
    \item \textbf{Wholebody 2D Pose Estimation}
    \begin{itemize}
        \item \textbf{HRNet}:\\
        Model type: Top-down;\\
        Configruation: W48;\\
        Epoch: 210;\\
        Dataset: COCO-Wholebody;\\
        Method: Heatmap;\\
        Optimizer: Adam;\\
        Learning rate: 5e-4;\\
        Warmup strategy: LinearLR;\\
        Scheduler: MultiStepLR;\\
        Input size: 384$\times$288.\\
    \end{itemize}

    \item \textbf{2D-to-3D Pose Lifting}:
    \begin{itemize}
        \item \textbf{JointFormer}:\\
        Epoch: 100;\\
        Dataset: H3WB;\\
        Learning rate: 1e-3;\\
        Learning rate decay: 1e5;\\
        Batch size: 64;\\
        Worker number: 8;\\
        Embedding type: Conv.\\
    \end{itemize}

    \item \textbf{Action Recognition}:
    \begin{itemize}
        \item \textbf{VideoMAE}:\\
        Model type: Recognizer3D;\\
        Backbone: VisionTransformer;\\
        Frame sampling strategy: 16$\times$4$\times$1;\\
        Epoch: 150;\\
        FLOPs: 180G;\\
        Dataset type: Kinetics-400;\\
        Input size: 224$\times$224.\\

        \item \textbf{TANet}:\\
        Method: RGB;\\
        Backbone: ResNet50;\\
        Pretrain: ImageNet;\\
        Frame sampling strategy: dense-1$\times$1$\times$8;\\
        Epoch: 150;\\
        FLOPs: 43.0G;\\
        Dataset type: Kinetics-400;\\
        Input size: 224$\times$224.\\

        \item \textbf{UniFormer}:\\
        Model type: Recognizer3D;\\
        Method: RGB;\\
        Backbone: UniFormer-B;\\
        Frame sampling strategy: 32$\times$4$\times$1;\\
        Epoch: 150;\\
        FLOPs: 59G;\\
        Dataset type: Kinetics-400;\\
        Input size: 320$\times$320.\\

        \item \textbf{TIN}:\\
        Method: RGB;\\
        Backbone: ResNet50;\\
        Pretrain: TSM-Kinetics400;\\
        Frame sampling strategy: 1$\times$1$\times$8;\\
        Epoch: 150;\\
        Dataset type: Kinetics-400;\\
        Input size: 256$\times$256.\\

        \item \textbf{TSN}:\\
        Method: RGB;\\
        Backbone: ResNet50;\\
        Pretrain: ImageNet;\\
        Frame sampling strategy: 1$\times$1$\times$8;\\
        Scheduler: MultiStep;\\
        Epoch: 150;\\
        FLOPs: 102.7G;\\
        Dataset type: Kinetics-400;\\
        Input size: 224$\times$224.\\

        \item \textbf{VideoSwin}:\\
        Method: RGB;\\
        Backbone: Swin-T;\\
        Pretrain: ImageNet-1K;\\
        Frame sampling strategy: 32$\times$2$\times$1;\\
        Epoch: 150;\\
        FLOPs: 88G;\\
        Dataset type: Kinetics-400;\\
        Input size: 224$\times$224.\\

        \item \textbf{TPN}:\\
        Method: RGB;\\
        Backbone: ResNet50;\\
        Pretrain: ImageNet;\\
        Frame sampling strategy: 8$\times$8$\times$1;\\
        Epoch: 150;\\
        Testing protocol: 10 clips$\times$3 crop;\\
        Dataset type: Kinetics-400;\\
        Input size: 320$\times$320.\\

        \item \textbf{X3D}:\\
        Method: RGB;\\
        Backbone: X3D-M;\\
        Frame sampling strategy: 13$\times$6$\times$1;\\
        Epoch: 100;\\
        Dataset type: Kinetics-400;\\
        Input size: 224$\times$224.\\

        \item \textbf{I3D}:\\
        Method: RGB;\\
        Backbone: ResNet50;\\
        Pretrain: ImageNet;\\
        Frame sampling strategy: 32$\times$2$\times$1;\\
        Epoch: 100;\\
        FLOPs: 43.5G;\\
        Dataset type: Kinetics-400;\\
        Input size: 224$\times$224.\\

        \item \textbf{I3D NL}:\\
        Method: RGB;\\
        Backbone: ResNet50 (NonLocalEmbedGauss);\\
        Pretrain: ImageNet;\\
        Frame sampling strategy: 32$\times$2$\times$1;\\
        Epoch: 100;\\
        FLOPs: 59.3G;\\
        Dataset type: Kinetics-400;\\
        Input size: 224$\times$224.\\
    \end{itemize}
\end{itemize}

\subsection{Datasets}
This section offers in-depth configurations for the NTU and H36M-based datasets that were utilized in our evaluation processes.

\subsubsection{NTU-based datasets}
Table~\ref{table:NTU-conf} presents the number of videos and frames for each dataset based on NTU.
\begin{table}
\begin{center}
\resizebox{0.5\textwidth}{!}{
\begin{tabular}{c c c}
\toprule[2pt]
\textbf{Dataset} & \textbf{Video Number} & \textbf{Frame Number} \\
\toprule[1pt]
\textbf{NTU-Original} & 1,911 & 173,129 \\
\textbf{NTU-GTAutoAct} & 18,792 & 953,570 \\
\bottomrule[2pt]
\end{tabular}
}
\end{center}
\caption{Configurations of NTU-based datasets.}
\label{table:NTU-conf}
\end{table}

\subsubsection{H36M-based datasets}
Table~\ref{table:H36M-conf} presents the number of videos and frames for each dataset based on H36M.
Table~\ref{class} presents the action classes of each H36M-based dataset.
\begin{table}
\begin{center}
\resizebox{0.5\textwidth}{!}{
\begin{tabular}{c c c}
\toprule[2pt]
\textbf{Dataset} & \textbf{Video Number} & \textbf{Frame Number} \\
\toprule[1pt]
\textbf{H36M-Original} & 480 & 1,302,398 \\
\textbf{H36M-Single} & 120 & 325,419 \\
\textbf{H36M-Extracted} & 432 & 80,000 \\
\textbf{H36M-SingleExtracted} & 108 & 20,000 \\
\textbf{H36M-GTAutoAct} & 3,384 & 872,564 \\
\bottomrule[2pt]
\end{tabular}
}
\end{center}
\caption{Configurations of H36M-based datasets.}
\label{table:H36M-conf}
\end{table}

\begin{table}
\begin{center}
\begin{tabular}{c  l}
\toprule[2pt]
\textbf{Label} ID & \textbf{Action}  \\
\toprule[1pt]
0 & Directions\\
1 & Discussion\\
2 & Eating\\
3 & Greeting\\
4 & Phoning\\
5 & Posing\\
6 & Purchases\\
7 & Sitting\\
8 & Sitting down\\
9 & Smoking\\
10 & Taking photo\\
11 & Waiting\\
12 & Walking \\
13 & Walking dog\\
13 & Walking together\\
\bottomrule[2pt]
\end{tabular}
\end{center}
\caption{Action classes in H36M-based datasets}
\label{class}
\end{table}

\section{Additional Results}
This section presents supplementary results from the experiments conducted in the main paper. 
It includes additional qualitative findings and detailed confusion matrices for a more comprehensive analysis.

\subsection{Additional Qualitative Results}
\begin{figure*}
\begin{center}
\includegraphics[width=\textwidth]{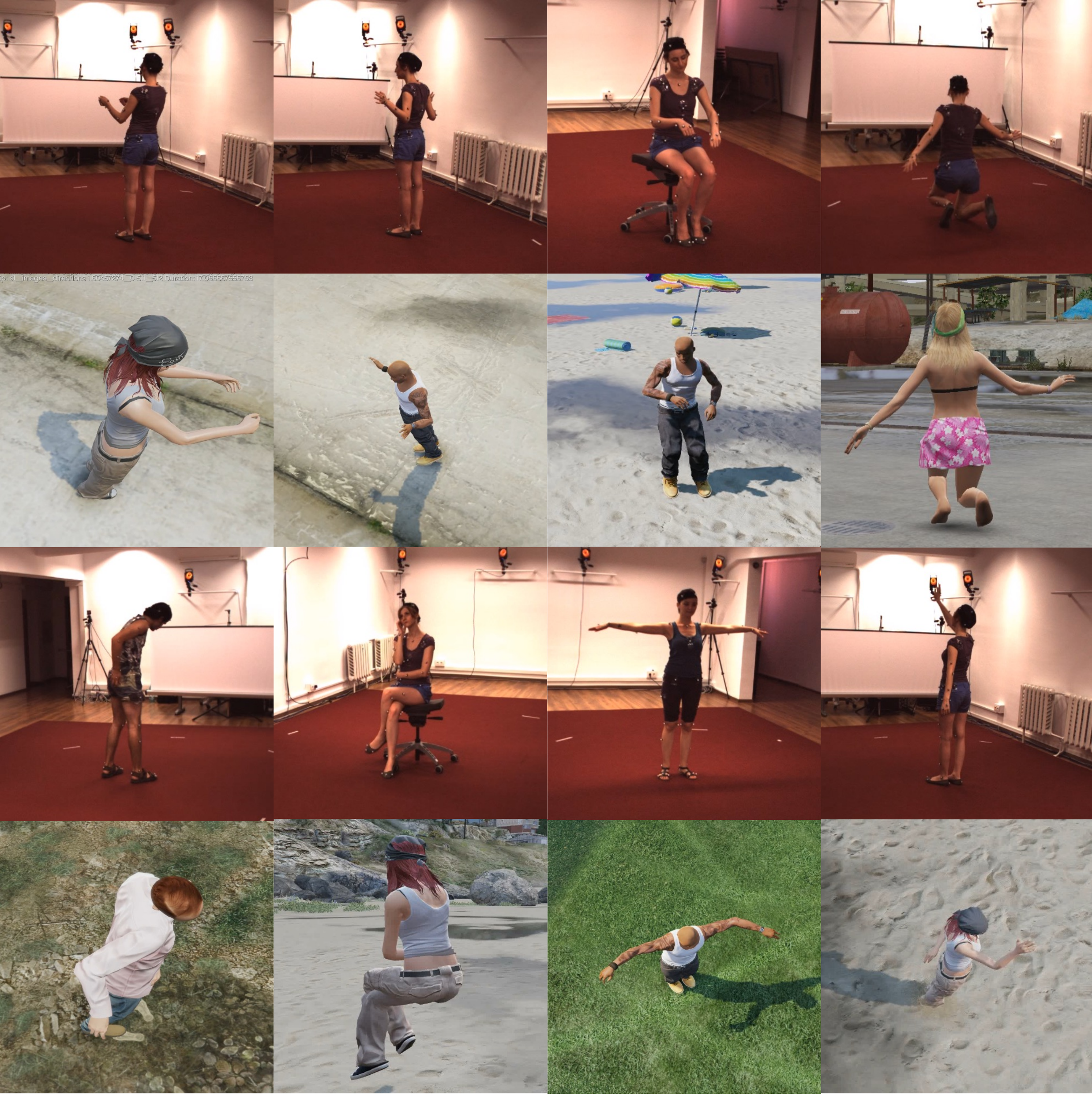}
\end{center}
   \caption{Additional qualitative results of whole-body representation by GTAutoAct, comparing with the original RGB frames in H36M.}
\label{add_res}
\end{figure*}

This section displays extra qualitative results of the whole-body representation created by GTAutoAct. 
These results are compared with the original RGB frames and are illustrated in Figure~\ref{add_res}.

\subsection{Confusion Matrices}
\begin{figure*}
\begin{center}
\includegraphics[width=\textwidth]{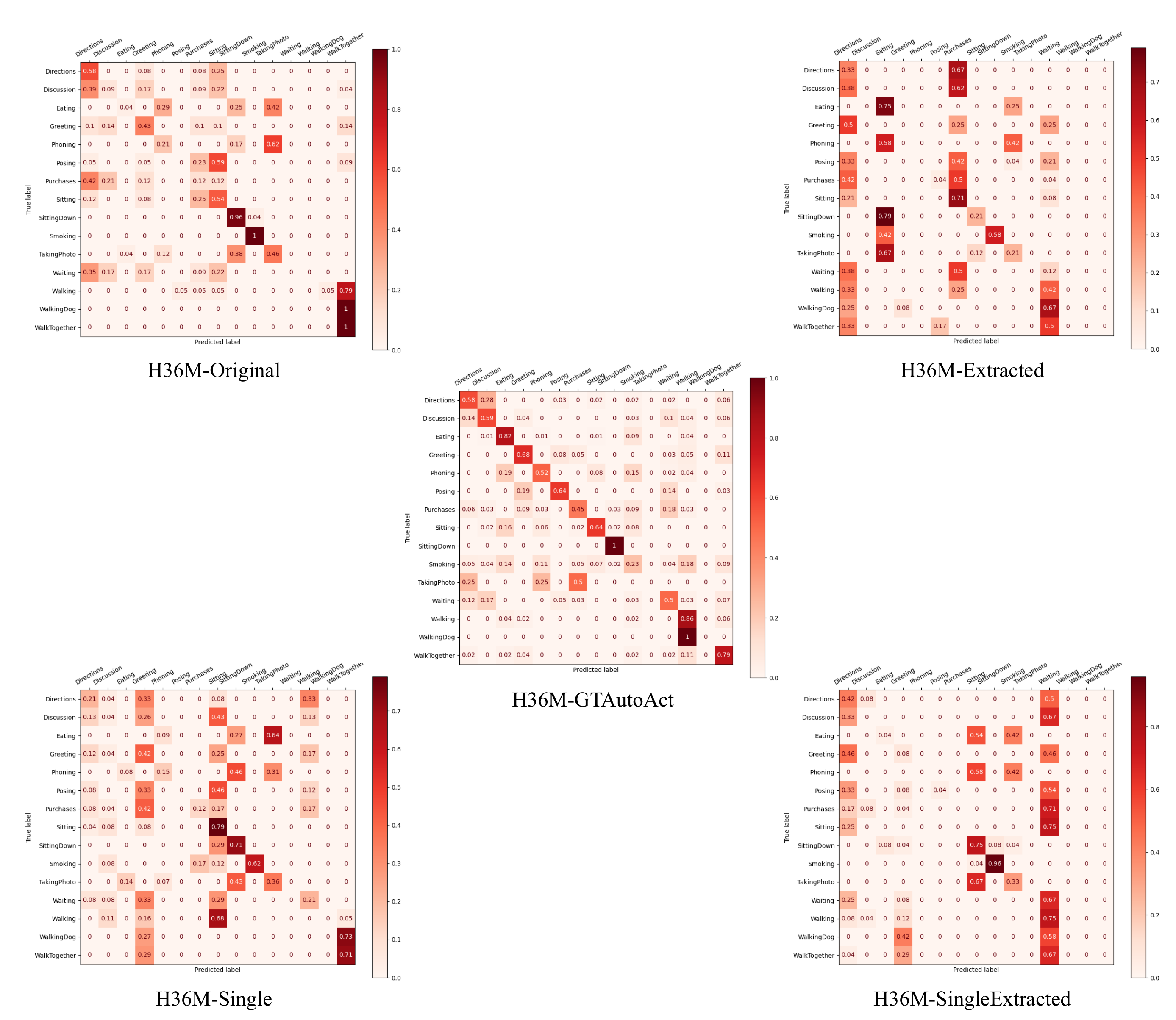}
\end{center}
   \caption{Confusion matrices of \textbf{TPN} model trained by each H36M-based datasets, tested by \textbf{H36M-Original-Test}.}
\label{cm_ori}
\end{figure*}

\begin{figure*}
\begin{center}
\includegraphics[width=\textwidth]{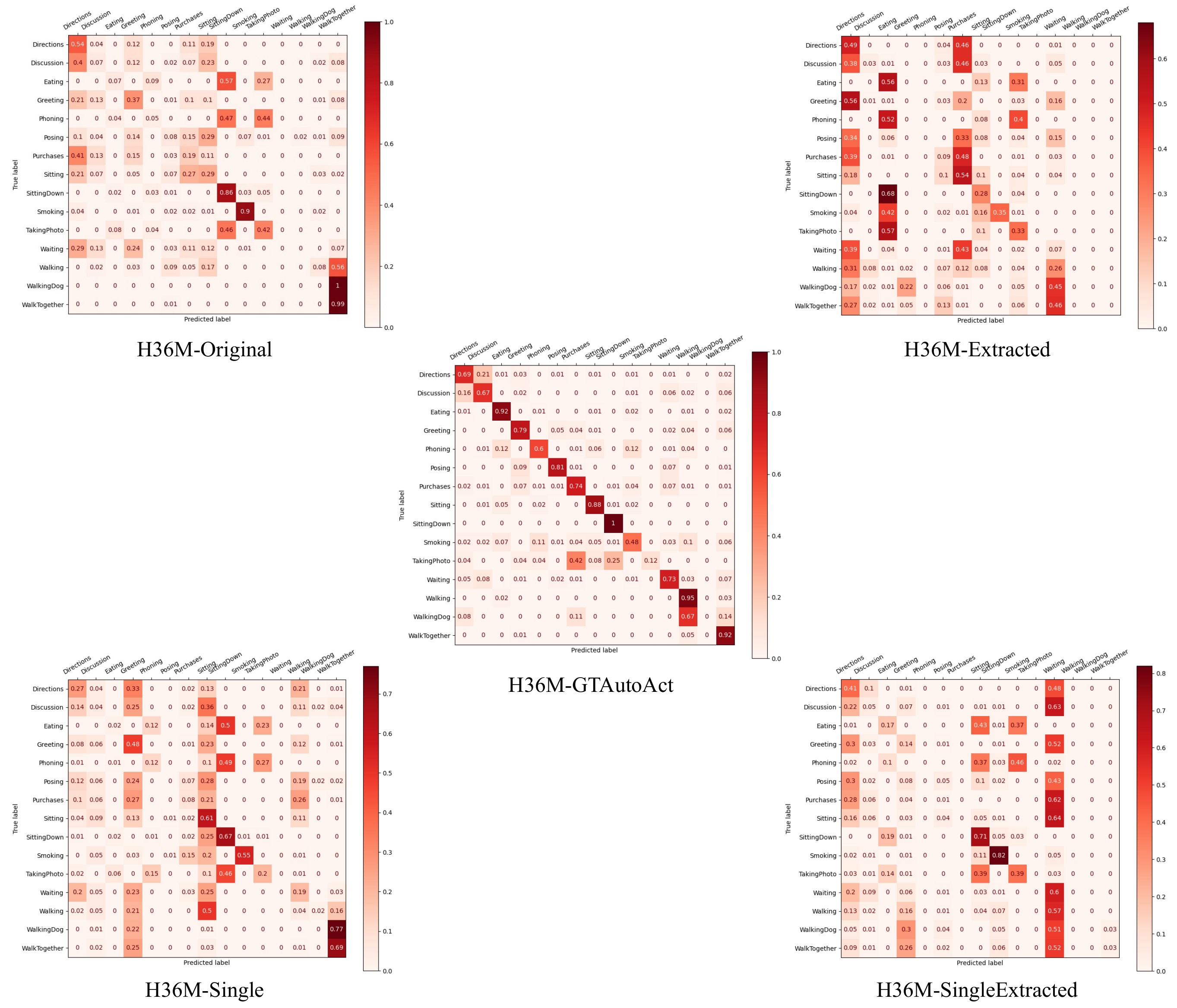}
\end{center}
   \caption{Confusion matrices of \textbf{TPN} model trained by each H36M-based datasets, tested by \textbf{H36M-Segment-Test}.}
\label{cm_seg}
\end{figure*}

This section includes selected confusion matrices for the TPN model trained on each H36M-based dataset. 
The performance of these models is evaluated using two different test datasets: H36M-Original-Test, with results displayed in Figure~\ref{cm_ori}, and H36M-Segment-Test, with results shown in Figure~\ref{cm_seg}.

\label{sec:rationale}

\end{document}